\documentclass{article} 
\usepackage[preprint]{colm2026_conference}

\usepackage{microtype}
\usepackage{hyperref}
\usepackage{url}
\usepackage{booktabs}

\usepackage{lineno}
\usepackage{hyperref}
\usepackage{url}
\usepackage{graphicx}
\usepackage{booktabs}
\usepackage{array}
\usepackage{subcaption}
\usepackage{enumitem}
\usepackage[table]{xcolor} 
\usepackage[most]{tcolorbox}
\usepackage{multirow}
\usepackage{float}
\usepackage{titlesec}
\usepackage{listings} 
\usepackage{placeins}
\usepackage{fvextra}

\usepackage{xcolor}

\definecolor{darkblue}{rgb}{0, 0, 0.5}
\hypersetup{colorlinks=true, citecolor=darkblue, linkcolor=darkblue, urlcolor=darkblue}

\title{How Do AI Agents Waste Your Money?}

\author{
  \hspace*{-0.175cm} \textbf{Huimin Zeng}\textsuperscript{1}\thanks{Equal contribution. Correspondence to huiminz3@illinois.edu and lu.yuxuan@northeastern.edu} \quad
  \textbf{Yuxuan Lu}\textsuperscript{2}\footnotemark[1] \quad
  \textbf{Jesse Cai}\textsuperscript{3} \quad
  \textbf{Eric Hanchen Jiang}\textsuperscript{4} \quad 
  \textbf{Di Wu}\textsuperscript{4} \quad  \\
  \textbf{Po-Nien Kung}\textsuperscript{4} \quad
  \textbf{Wenbo Hu}\textsuperscript{4} \quad
  \textbf{Ziyi Wang}\textsuperscript{2} \quad
  \textbf{Dakuo Wang}\textsuperscript{2} \quad 
  \textbf{Jiaxin Pei}\textsuperscript{5}\thanks{Equal advising.} \quad
  \textbf{Kai-Wei Chang}\textsuperscript{4}\footnotemark[2] \\
  \textsuperscript{1}University of Illinois Urbana-Champaign \quad
  \textsuperscript{2}Northeastern University \quad \\
  \textsuperscript{3}Independent Researcher \quad
  \textsuperscript{4}University of California, Los Angeles \quad
  \textsuperscript{5}UT Austin 
}

\begin{document}

\ifcolmsubmission
\linenumbers
\fi
\maketitle

\begin{abstract}
AI agents may consume massive tokens across a wide range of tasks, but how many of those tokens are really necessary? Although prior work has shown that agents can incur unexpectedly high token costs, we still lack a systematic way to identify and measure token waste\footnote{We differentiate ``token cost'' and ``token waste'' in this work, where token cost refers to the overall bill, while token waste refers to the tokens that are useless and removable in the agent's trajectory.}. In this work, we introduce a unified framework for measuring token waste across heterogeneous agent systems. Our framework records and normalizes complete agent trajectories, identifies removable spans using an LLM-as-a-Judge (LLMaJ), and maps the identified text spans back to the original trajectory for token-level measurement. We validate the identified waste through direct interventions by removing marked spans, resuming agent execution, and measuring changes in task outcomes and subsequent actions. In online pruning experiments, we remove 41-46\% of the model's reasoning while largely preserving task performance across two distinct agent benchmarks.
Using this framework, we analyze agents across multiple settings. Our analysis shows that token waste has a system-level structure that aggregate token usage alone cannot reveal. First, waste varies more across tasks and agent settings than across models. Second, reasoning and visible execution can exhibit different waste patterns. Third, different harnesses can introduce distinct sources of inefficiency even under the same model and task. Overall, our framework transforms token waste into an actionable optimization signal, enabling researchers and practitioners to train more efficient models, design better reasoning strategies, and optimize agent harnesses toward reducing unnecessary computation rather than merely reducing token usage.
\end{abstract}

\section{Introduction}
\label{sec:intro}
The recent emergence of AI agents has transformed the pattern of LLM use from single-turn QA-style interaction to multi-turn interaction with tools, environments, and the user \citep{yao2022react,liu2024agentbench}. Such multi-turn interaction is driven by the agent harness, which wraps the LLM and orchestrates and standardizes what the model sees, what information is retained in memory, which actions it can take, and how observations are returned to the model \citep{yang2024swe,wu2023autogen,packer2023memgpt,openai2025codex,wang2025openhands}. Through such orchestration, agents can handle complex and long-horizon tasks, but the resulting multi-turn execution is also particularly expensive \citep{ruan2026beyond,xu2026adapting,yao2026harness,luo2025ultrahorizon,bai2026ai}. For instance, a single trajectory may involve many rounds of reasoning, responses, tool calls, and observations, consuming substantially more tokens than a conventional LLM query.

However, token consumption and token waste are distinct quantities. For example, a long trajectory may be expensive because the task genuinely requires extensive reasoning and multiple turns of interaction, whereas a shorter trajectory may still spend many tokens on repeating information or redundant tool calls. Despite this distinction, existing studies on agent economics largely characterize agent efficiency with aggregated quantities such as token usage, latency, or overall monetary cost \citep{bai2026ai,satish2026beyond,lin2026bagen}. Such aggregate measurements quantify how much computation an agent uses, although they leave open how much of that computation was actually necessary. As a result, efforts to reduce agent cost based on these measurements can only target the total token count, which risks cutting the reasoning and interaction a task needs while leaving removable content in place.


\begin{figure*}[t]
    \centering
    \includegraphics[width=\textwidth]{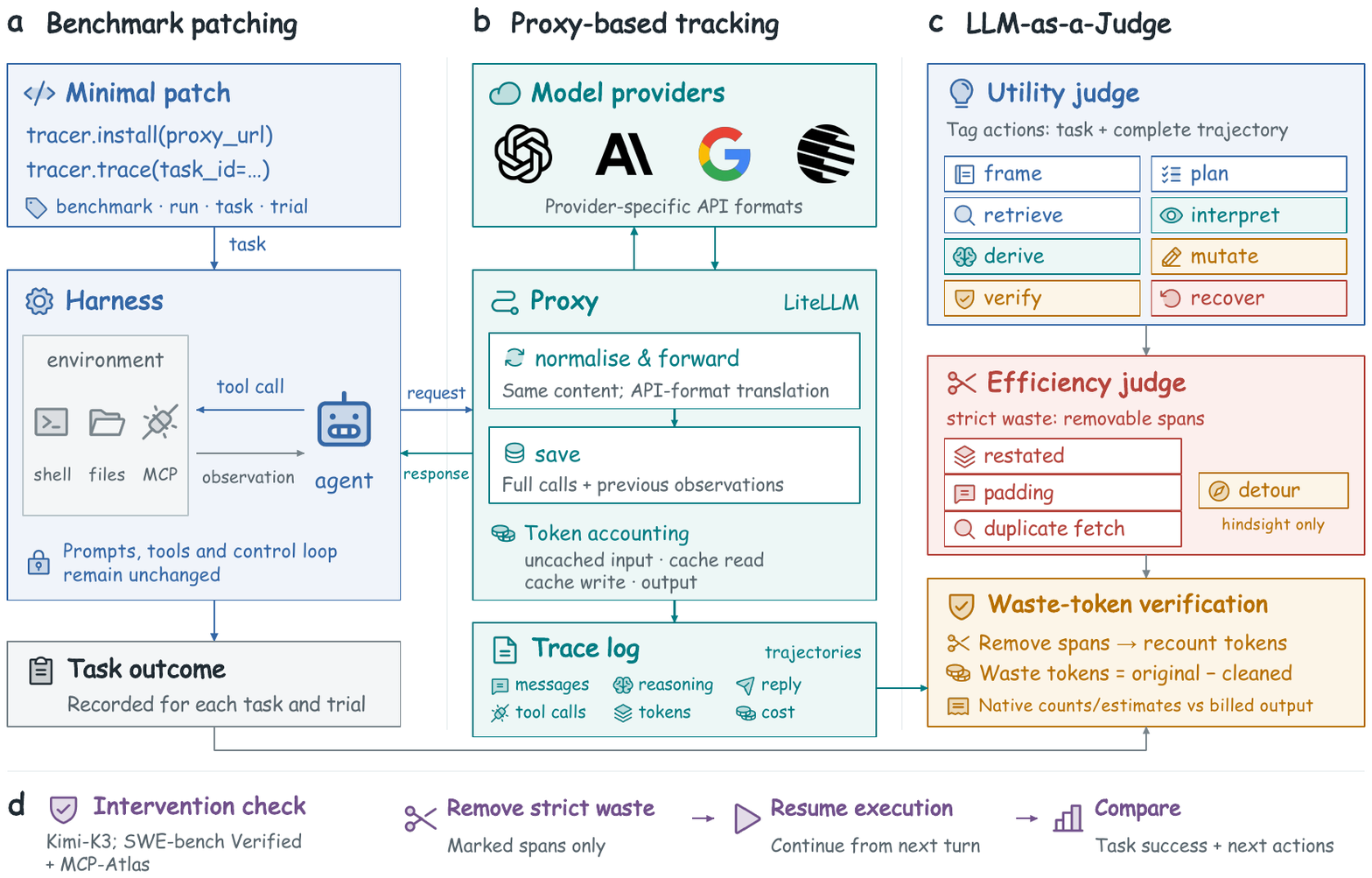}
    \caption{Overview of the proposed framework. The framework minimally instruments agent benchmarks, routes model interactions through a unified proxy for trajectory tracing, and uses LLM-based judges to annotate agent actions and identify token waste, followed by span-level verification and token measurement.}
    \label{fig:fig0_framework_overview}
\end{figure*}

The gap between consumption and waste therefore raises a fundamental measurement question: \textbf{how can we identify which parts of an agent trajectory are unnecessary and removable?} Answering this question is challenging for several reasons. First, the length of responses or trajectories alone is insufficient to determine waste, because text that appears verbose may still contribute to a later agent turn, while a seemingly reasonable exploratory step may turn out to be unnecessary in hindsight. Second, agent trajectories are heterogeneous, since different models and harnesses expose reasoning, responses, tool calls, and token accounting in different formats, which makes consistent measurement across systems difficult. Third, identifying apparently unnecessary text is insufficient on its own, given that a meaningful notion of waste should be supported by evidence that removing the identified content does not significantly disrupt subsequent agent execution or task outcomes.

To address these challenges, we introduce a unified framework for \textbf{measuring token waste in AI agents}. As shown in Figure~\ref{fig:fig0_framework_overview}, the framework consists of three stages, namely tracing, identifying, and verifying. First, heterogeneous benchmarks and harnesses can be connected to the framework with only a few lines of code changes, and the resulting agent trajectories are normalized into a common representation across models, benchmarks, and harnesses. Second, token waste is defined as content that can be removed without eliminating useful work from the trajectory, and based on this definition, an LLM-as-a-Judge (LLMaJ) identifies such waste spans within reasoning, visible responses, and tool interactions. The framework further distinguishes directly removable \textbf{strict waste} from exploratory \textbf{detours}, which may be unnecessary in hindsight but reasonable during agent execution. Finally, the identified waste is verified and quantified using token-counting tools provided by the original model providers. As a result, we can measure wasted tokens and monetary cost under each provider's native accounting scheme while applying the same span-level definition of waste across heterogeneous agent systems.

Before using the framework for large-scale analysis, we validate whether the identified waste is consistent with our removability criterion. Specifically, the validation experiments directly intervene on agent trajectories by removing identified waste and measuring its effect on final task outcomes and subsequent actions. In addition, we prune identified waste online while the agent is running. Across these validation experiments, removing identified waste generally preserves task performance and produces limited behavioral changes. In online pruning experiments, we remove 41-46\% of the model's reasoning while largely preserving task performance across SWE-bench Verified and MCP-Atlas.

Using our framework, we investigate token waste across multiple LLMs, tasks, reasoning-effort settings, and agent harnesses. In doing so, the analysis moves beyond traditional token usage metrics and examines where unnecessary computation arises within an agent system. Our results suggest that several components of the agent system beyond the underlying LLM shape token waste. Moreover, reasoning and visible execution can exhibit different waste patterns, and different harnesses can introduce distinct sources of inefficiency even when the model and task are fixed. Together, these analyses demonstrate how our framework can detect and attribute token waste in agent trajectories that would otherwise be hidden by traditional token usage metrics.

Our contributions are summarized as follows:
\begin{itemize}
    \item We introduce a unified framework for measuring token waste in heterogeneous AI-agent systems. The framework distinguishes \textbf{strict waste} from exploratory \textbf{detours} and validates the identified waste through direct interventions on agent trajectories.

    \item We apply the framework across models, tasks, reasoning-effort settings, and agent harnesses, showing how it can localize and attribute inefficiencies that traditional aggregate metrics cannot reveal. The resulting waste annotations can also provide supervision and diagnostic signals for training more token-efficient models and optimizing agent harnesses.



    \item We open-source our framework, code, and labeled agent trajectories to support future research on token-efficient agent systems.\footnote{Code is available at \url{https://github.com/huiminzeng/How_Do_AI_Agents_Waste_Your_Money}, and data are available at \url{https://huggingface.co/datasets/huiminz3/How_Do_AI_Agents_Waste_Your_Money}.}
\end{itemize}
\section{Token-Waste Analysis Framework}
\label{sec:method}
\subsection{Unified Agent Tracking}
Agent systems expose execution traces in heterogeneous formats across models, providers, benchmarks, and harnesses. To analyze token waste consistently across these systems, the first stage of our framework uses a unified tracking framework to capture complete agent executions and normalize them into a common representation. By design, the tracking framework supports arbitrary API endpoints and requires only minimally invasive changes to existing agent harnesses and benchmark code.

To define what the tracking framework captures, we follow prior work~\citep{yao2022react,wang2025steca} and represent an agent trajectory as a sequence of turns. At each turn, the harness sends the current context to the model, which returns its reasoning, a visible response, and tool calls, while the provider reports the token usage of that invocation. Depending on the model provider, the reasoning may be fully exposed (e.g., Kimi-K3), summarized (e.g., Gemini 3.1 Pro), or unavailable (Claude models). After a tool is executed, its result is returned to the harness as a new observation and appended to the context, and the interaction continues until the agent terminates or reaches its execution limit.

To keep the changes to existing agent code minimally invasive, the tracking framework attaches to a harness through its standard model API (Figure~\ref{fig:fig0_framework_overview}). Specifically, the harness's API base URL is set to our proxy, so its model requests reach the proxy through the harness's own client code. To associate each request with its task, a small patch installs a tracer and wraps each evaluation unit in a trace scope that labels it with identifiers such as the benchmark, run, task, and trial. Concretely, the tracer patches the common Python HTTP client libraries, so every request sent to the proxy carries the current trace scope in a request header. For harnesses that cannot add request headers, such as OpenClaw, each evaluation unit instead runs with its own proxy instance, which receives the unit's trace scope at startup. In both cases, the harness's prompts, tools, and control loop run unchanged, so existing agent code can be traced without being rewritten.

The proxy is built on LiteLLM\footnote{\url{https://github.com/BerriAI/litellm}}, which routes each requested model name to its configured backend endpoint and translates between provider-specific API formats, so the tracking framework stays independent of the API endpoint. On the request side, the proxy forwards each request to the backend provider with its semantic content unchanged. Meanwhile, the proxy records the trace scope of each request for logging and trajectory storage, without forwarding it to the provider. On the response side, a logging callback records each call together with its full request messages, the response content and tool calls, and the token usage, which it normalizes across providers into uncached input, cache-read, cache-write, and output tokens. Since each request carries the observations returned after the previous call, the logged calls together contain the complete interaction. As a result, the harness continues to execute normally while the proxy captures every turn, and supporting a new API endpoint only requires adding it to the proxy configuration.

\begin{figure*}[t]
    \centering
    \includegraphics[trim=0cm 1.5cm 0cm 0cm, clip, width=\textwidth]{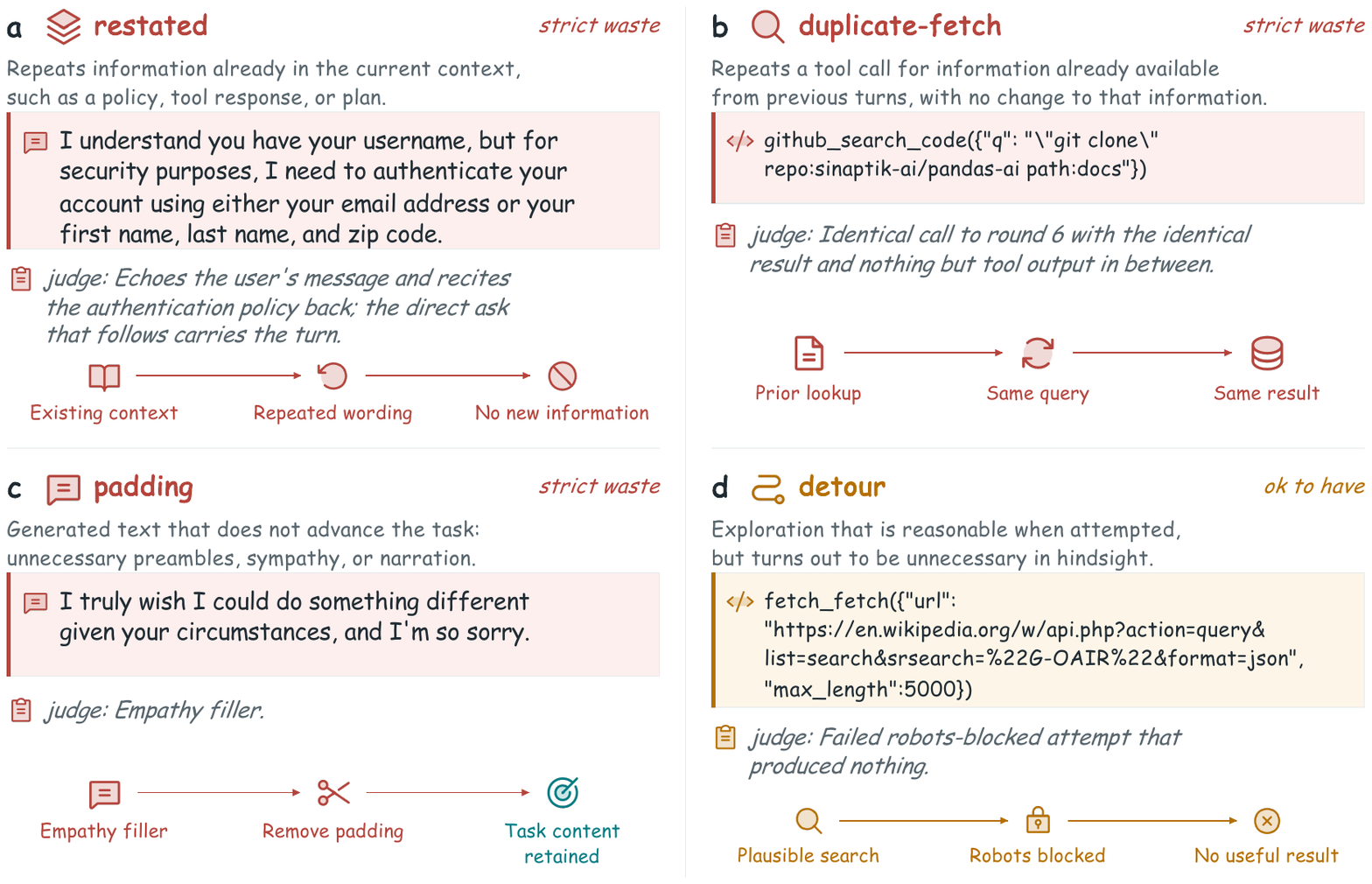}
    \caption{Examples of the waste patterns drawn from real agent trajectories.}
    \label{fig:waste_kinds_examples}
\end{figure*}

\subsection{Token Waste Patterns}
Our definition of token waste follows a principle of \textbf{minimalism}, under which an agent trajectory should contain no more generated content than is necessary to preserve the useful work required for completing the task. Accordingly, we treat content as waste when it can be removed without eliminating useful work from the trajectory. Under this criterion, the goal is to identify as much removable content as possible while preserving the task-relevant computation and execution.

To characterize how such waste appears in practice, we manually inspect a subset of agent trajectories and identify four recurring patterns, namely \textbf{Restate}, \textbf{Duplicate-Fetch}, \textbf{Padding}, and \textbf{Detour}.

\paragraph{Restate} refers to cases where the agent repeats information that already exists in the current context. For example, the agent may quote a policy document that already appears in the system prompt, paraphrase a tool response that remains available in the context, or repeatedly restate the same plan before each step.

\paragraph{Duplicate-Fetch} refers to repeated tool calls that retrieve information already available from previous turns and that has not changed. Examples include reading the same file multiple times or repeating the same lookup against an unchanged database.

\paragraph{Padding} refers to generated text that does not help advance the task, such as unnecessary preambles, sign-offs, restatements of the user's request, or narration of an action immediately before performing it. For example, in a customer-service task, an agent may generate an extended expression of sympathy that does not contribute to resolving the user's request.

\paragraph{Detour} refers to exploratory work that appears reasonable when attempted but ultimately turns out to be unnecessary for completing the task. Examples include exploring a plausible approach that leads nowhere or retrieving information that is never used later in the trajectory.

A further distinction is made between \textbf{strict waste} and \textbf{detours}. Specifically, Restate, Duplicate-Fetch, and Padding are treated as strict waste because they can generally be removed without eliminating useful work. In contrast, detours are treated separately because exploration is sometimes necessary. After all, an exploratory path may be reasonable given the information available at that time step, even if the task could have been completed without it later. In this sense, strict waste is directly avoidable, whereas detours are unnecessary in hindsight and could potentially be reduced through better optimization. For a more concrete illustration of these waste patterns, Figure~\ref{fig:waste_kinds_examples} shows some real examples drawn from the trajectories collected in our experiments.

\subsection{LLM-as-a-Judge Waste Identification}
To identify token waste at scale, we use an LLM as a Judge (LLMaJ) to analyze each collected trajectory. The judge also follows the minimalism principle as we define waste patterns. Given the task and the complete trajectory, it identifies content that can be removed without eliminating useful work needed to complete the task. The judge's prompt also provides explicit criteria for distinguishing among the four waste patterns. The judge then marks the corresponding spans and assigns each marked span to one of the four categories.

After the judge has marked a trajectory, our framework removes the identified spans and recounts the remaining tokens. The token difference between the original and the cleaned trajectory therefore gives the number of wasted tokens. From these counts, the framework then computes the corresponding monetary cost and waste ratio.

As a consistency check, the marked tokens are additionally counted or estimated using the tokenizers or token-counting tools available from each model provider. The total count reconstructed in this way is then compared against the output-token usage reported by the provider. In our experiments, the reconstructed counts closely match the provider-reported usage, apart from small provider-specific formatting differences. As a result, we can reliably estimate the number of wasted tokens and their corresponding monetary cost under each provider's native token-accounting scheme.

\section{Experimental Setup}
Our experiments cover multiple frontier LLMs, benchmarks, reasoning settings, and agent harnesses. The benchmarks include \textbf{$\tau^2$-bench \citep{barres2025tau2}} (airline and retail), \textbf{SWE-bench Verified} \citep{jimenez2024swebench}, \textbf{Claw-Eval} \citep{ye2026clawevaltrustworthyevaluationautonomous} and \textbf{MCP-Atlas} \citep{bandi2026mcp}. These benchmarks span customer service, software engineering, and general-purpose tool-use scenarios with different interaction patterns and environments. 

Six frontier LLMs are considered: \textbf{Kimi-K3} \citep{team2026kimi}, \textbf{Gemini 3.1 Pro} \citep{google2026gemini31pro}, \textbf{GPT-5.6 Terra} \citep{openai2026gpt56}, \textbf{GPT-6 Astra} \citep{openai2026gpt6astra}, \textbf{Claude Sonnet 5} \citep{anthropic2026sonnet5}, and \textbf{Claude Fable 5.1} \citep{anthropic2026fable51}.

\paragraph{Validation.} Section~\ref{sec:sanity_check} evaluates whether the token waste identified by our framework is removable through direct interventions on agent trajectories.

\paragraph{Models and Tasks.}  {Section~\ref{sec:how_many_tokens_is_your_AI_agent_wasting}} compares token waste across models and tasks using each benchmark's native harness. 

\paragraph{Reasoning Effort.} In {Section~\ref{sec:when_more_reasoning_stops_paying_off}}, we vary the reasoning effort of Kimi-K3 on the same tasks to study how additional reasoning affects agent behavior and token waste. 

\paragraph{Agent harnesses.} {Section~\ref{sec:same_model_different_agent}} compares five different agent harnesses (i.e., \textbf{OpenCode}~\citep{opencode2026}, \textbf{OpenClaw}~\citep{openclaw2026}, \textbf{OpenHands}~\citep{wang2025openhands}, \textbf{Hermes}~\citep{nous2026hermes}, and \textbf{Kimi Code CLI}~\citep{moonshot2026kimicode}) using Kimi-K3 as the backend LLM. This controlled comparison demonstrates how harness design affects agent behavior and token waste.
\section{Validating Token-Waste Identification}
\label{sec:sanity_check}

LLMaJ waste identification is evaluated through direct interventions on agent trajectories. The key question is whether spans identified as strict waste can be removed without significantly changing the agent's subsequent behavior or final task outcome. To test this, we directly intervene on agent trajectories by removing identified spans and measuring their effects on benchmark success and subsequent actions.

\subsection{Intervention Setup}
For the intervention-based validation, we focus on Kimi-K3 trajectories from SWE-bench Verified and MCP-Atlas. Kimi-K3 is selected because its reasoning traces are fully exposed, allowing us to directly remove reasoning spans identified as waste. SWE-bench Verified and MCP-Atlas represent two different agent settings: software-engineering tasks with a stateful execution environment and general-purpose tool-use tasks with comparatively lightweight environment state. In both benchmarks, reasoning from previous turns is retained in the model's subsequent context.

\subsection{Does Removing Identified Waste Change Task Outcomes?}
The first experiment tests whether removing identified waste changes the final task outcome. For each trajectory, we select an intervention point at different stages of execution, remove the strict-waste spans that appeared before this point, and resume execution from the next turn. As shown in Table~\ref{tab:correctness-check}, removing LLMaJ-identified strict waste generally preserves task performance, with the first-turn intervention on MCP-Atlas being the main exception.

\begin{table}[!bht]
\centering
\small
\begin{tabular}{llrrrcr}
\toprule
 & & \multicolumn{2}{c}{removed} & \multicolumn{2}{c}{solved (\%)} & \\
\cmidrule(lr){3-4}\cmidrule(lr){5-6}
Benchmark & Intervention & median & \% of & original & pruned & worse / \\
 & & tokens & prefix & & & better \\
\midrule
SWE-bench Verified  & 25\%         & 350 & 19.8 & 81.7 & \textbf{81.7} & 3 / 3 \\
                    & 50\%         & 649 & 14.2 & 86.0 & \textbf{86.0} & 1 / 1 \\
                    & 75\%         & 813 & 15.9 & 91.3 & \textbf{91.3} & 0 / 0 \\
\addlinespace
MCP-Atlas           & 1st (turn 1) & 111 & 47.5 & 75.3 & \textbf{65.6} & 9 / 0 \\
                    & 2nd          & 163 & 40.8 & 74.2 & \textbf{74.2} & 2 / 2 \\
                    & 3rd          & 281 & 32.0 & 71.8 & \textbf{73.2} & 0 / 1 \\
\bottomrule
\end{tabular}
\caption{
Effect of removing LLMaJ-identified strict waste and resuming agent execution.
``\% of prefix'' denotes the fraction of model-generated text that is removed, excluding tool observations. ``Worse / better'' counts units that change from solved to unsolved and vice versa. For MCP-Atlas, the trajectories are shorter, so we only intervene after the first, second, and third turns.}
\label{tab:correctness-check}
\end{table}

\subsection{Does Removing Identified Waste Change the Next Action?}
The next experiment tests whether LLMaJ-identified waste affects the agent's subsequent action. For each sampled turn, marked spans are removed from the reasoning of previous turns, and the model generates a fresh next turn. We sample 16 continuations from both the original and intervened contexts and compare the resulting action distributions.

An action is represented by its tool call, including the tool name and arguments, or by the visible reply when no tool is called. Given this representation, the difference between two actions \(a\) and \(b\) is measured using Jaccard distance over their token sets:
\[
d(a,b)
=
1 -
\frac{|T(a)\cap T(b)|}
{|T(a)\cup T(b)|}.
\]

As such, for two sets of sampled actions \(A\) and \(B\), the shift between their distributions is measured using energy distance:
\[
\mathcal{E}(A,B)
=
\frac{2}{|A||B|}
\sum_{a\in A}\sum_{b\in B} d(a,b)
-
\frac{1}{\binom{|A|}{2}}
\sum_{a<a'} d(a,a')
-
\frac{1}{\binom{|B|}{2}}
\sum_{b<b'} d(b,b').
\]

For each turn, significance is estimated using a permutation test with 2,000 random reassignments of the 32 sampled actions. We report the fraction of turns in which removing waste from previous reasoning produces a significant shift in the next-action distribution. As a positive control, the same experiment is repeated after removing all previous reasoning.

\begin{table}[!bht]
\centering
\small
\begin{tabular}{llr}
\toprule
Benchmark & Intervention & Changed turns (\%) \\
\midrule
SWE-bench Verified
    & Marked waste deleted  & \textbf{3.0} \\
    & All reasoning deleted & \textbf{25.0} \\
\addlinespace
MCP-Atlas
    & Marked waste deleted  & \textbf{10.5} \\
    & All reasoning deleted & \textbf{37.9} \\
\bottomrule
\end{tabular}
\caption{
Effect of removing previous reasoning on the agent's next-action distribution.
Each turn is evaluated using 16 rollouts per condition and an energy-distance permutation test.
``Changed turns'' counts turns for which the intervention produces a significant shift in the action distribution at $p<0.05$.
}
\label{tab:correctness-action}
\end{table}

As shown in Table~\ref{tab:correctness-action}, removing LLMaJ-identified waste from earlier reasoning has only a limited effect on the agent's next-action distribution. On SWE-bench Verified, only 3.0\% of turns show a significant shift, compared with 25.0\% when the entire previous reasoning history is removed. When the intervention effects are aggregated across turns, the shift from removing marked waste is statistically detectable ($p=0.007$), but its magnitude is only 1.8\% of the natural spread among the control rollouts. Thus, the measurable effect is very small relative to the model's own variation across samples. On MCP-Atlas, removing marked waste changes the action distribution in 10.5\% of turns, compared with 37.9\% after removing all previous reasoning. The effect of marked-waste removal is therefore more pronounced than on SWE-bench Verified, but remains substantially smaller than the effect of removing the reasoning history as a whole.

\subsection{Pruning Waste During Agent Execution}
The final intervention tests whether identified waste can be removed online, while the agent is actively solving the task. At each turn, an online judge examines the current reasoning together with the preceding context and identifies waste. The online judge is restricted to \textbf{Restate} and \textbf{Padding}, because these patterns can be identified from the current context without observing future turns. In contrast, identifying \textbf{Detours} may require information from later in the trajectory, while \textbf{Duplicate-Fetch} concerns redundant tool calls rather than reasoning spans. The marked spans are removed, and the model regenerates its response and tool call conditioned on the pruned reasoning. The pruned reasoning and regenerated action are then added to the context for subsequent turns.

\begin{table}[!bht]
\centering\small
\begin{tabular}{@{}lrrrr@{}}
\toprule
& \multicolumn{2}{c}{SWE-bench Verified} & \multicolumn{2}{c}{MCP-Atlas} \\
\cmidrule(lr){2-3}\cmidrule(lr){4-5}
Pruning Method & deleted (\%) & solved (\%) & deleted (\%) & solved (\%) \\
\midrule
nothing deleted & 0 & \textbf{89.0} & 0 & \textbf{65.0} \\
restated and padding deleted & 41 & \textbf{89.5} & 46 & \textbf{62.3} \\
all reasoning deleted & 100 & \textbf{73.0} & 100 & \textbf{56.0} \\
\bottomrule
\end{tabular}
\caption{Online pruning results on SWE-bench Verified and MCP-Atlas.}
\label{tab:cc-online}
\end{table}

As shown in Table~\ref{tab:cc-online}, pruning LLMaJ-identified waste has much less impact on task success than removing the reasoning entirely. This pattern holds across both SWE-bench Verified and MCP-Atlas, suggesting that the identified spans can be removed during agent execution with relatively limited effect on performance.

To summarize, these intervention experiments provide complementary evidence for our waste identification. The results support both our minimalism-based definition of token waste and the effectiveness of LLMaJ in identifying content that can be removed with limited impact on agent behavior and task performance.
\section{How Many Tokens Is Your AI Agent Wasting?}
\label{sec:how_many_tokens_is_your_AI_agent_wasting}
With the framework established and validated, we next apply it across models and tasks to measure the prevalence and composition of token waste in agent trajectories. Across the collected trajectories, roughly 18\% of output tokens are identified as waste. Strikingly, for more than 45\% of tasks, the wasted output of one agent alone exceeds the entire output of another agent that successfully solves the same task.

\subsection{How Much of the Output Bill Is Wasted?}
For the cross-model comparison, we measure the fraction of each model's output cost attributable to identified waste. The analysis covers visible responses and tool calls across all models. For Kimi-K3, which also exposes its full reasoning trace, reasoning waste is reported separately. Results are averaged across the four benchmarks.

\begin{figure*}[!bht]
    \centering
    \includegraphics[trim=0cm 0cm 0cm 0cm, clip, width=0.96\textwidth]{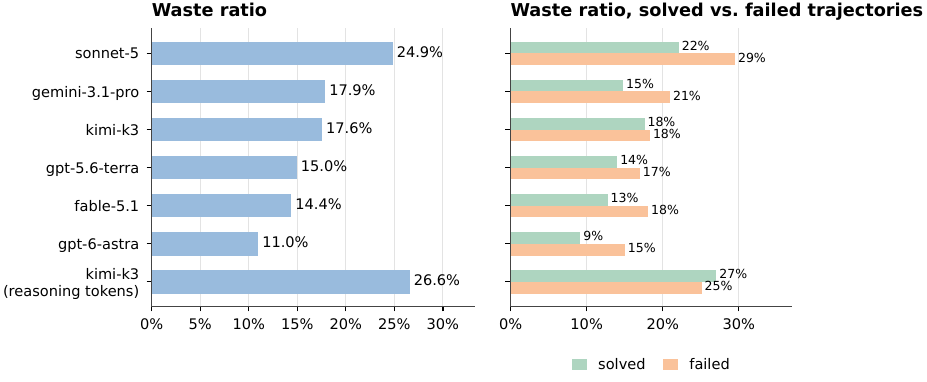}
    \caption{Averaged waste ratio comparison across the evaluated models. Waste ratio is calculated by dividing the cost of wasted output by the cost of all of its audited output, and we average this waste ratio across the four benchmarks.} 
    \label{fig:fig2_waste_by_model}
\end{figure*}

Figure~\ref{fig:fig2_waste_by_model} shows that the waste ratio varies substantially across models. For all six models, failed trajectories generally have higher waste ratios in visible responses and tool calls than successful trajectories. Kimi-K3 shows the opposite pattern in its reasoning, where successful trajectories have a higher waste ratio. These contrasting patterns suggest that token waste is not simply a consequence of task failure.

\begin{figure*}[!bht]
    \centering
    \includegraphics[width=\textwidth]{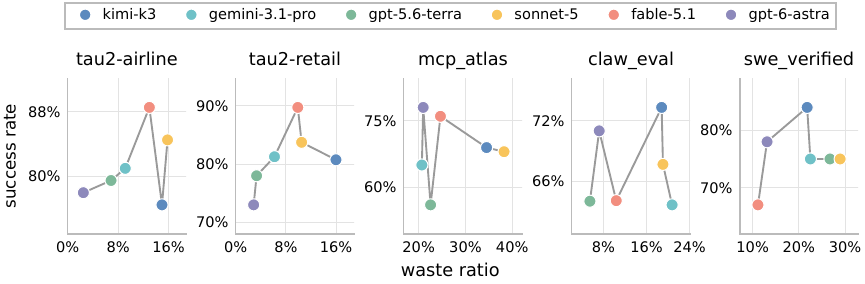}
    \caption{Performance-Waste Pareto shows that models with less waste can still achieve higher success rates than models with more waste.}
    \label{fig:fig3_waste_vs_success}
\end{figure*}

Given that failure alone does not explain waste, the next question is whether higher waste is associated with better task performance. Across models and benchmarks, no clear trade-off emerges between waste ratio and success rate. As shown in Figure~\ref{fig:fig3_waste_vs_success}, in several benchmarks, models with less waste can still achieve higher success rates than models with more waste. These results suggest that agents can reduce waste without giving up performance. 

\begin{figure*}[!bht]
    \centering
    \includegraphics[width=\textwidth]{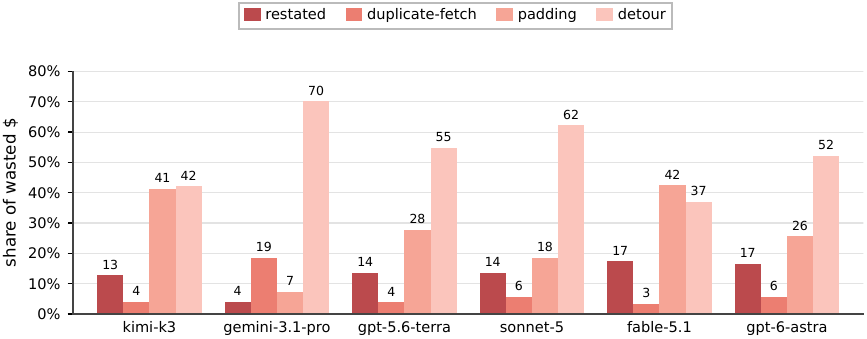}
    \caption{Waste pattern distribution for different models.  Although the distributions differ, detours dominate the waste across almost all of them.}
    \label{fig:fig4_waste_by_kind}
\end{figure*}

If waste does not buy success, what do the wasted tokens actually contain? Section~\ref{sec:method} defines four systematic waste patterns: \textbf{Restate}, \textbf{Duplicate-Fetch}, \textbf{Padding}, and \textbf{Detour}. Figure~\ref{fig:fig4_waste_by_kind} shows the distribution of these waste patterns across models. Although the distributions differ, detours generally dominate the waste across almost all of them. This is expected, because the agents have to explore unfamiliar tools and environments. In these cases, the agents often need to try several paths before they find one that works, which can lead to potentially wasteful detours. These results suggest that reducing unnecessary exploration may be an important direction for improving agent efficiency. We hypothesize that potential approaches include better tool selection and routing, more structured representations of tools and environments, and reusable knowledge that helps agents avoid repeatedly exploring similar paths.

\subsection{Which Agentic Tasks Waste the Most?}
\begin{figure*}[t]
    \centering
    \includegraphics[width=\textwidth]{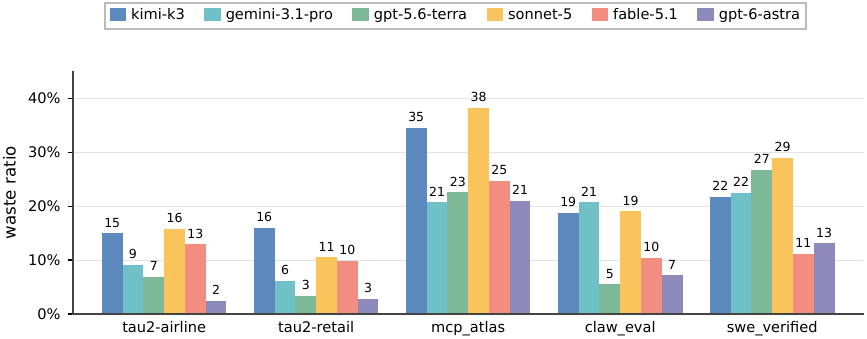}
    \caption{Comparing waste ratio across the benchmarks, waste ratios vary much more across benchmarks than across model.}
    \label{fig:fig5_waste_by_benchmark}
\end{figure*}

Exploration depends on the environment, so we next break the waste down by benchmark. Figure~\ref{fig:fig5_waste_by_benchmark} demonstrates that waste varies much more across benchmarks than across models. Among the evaluated benchmarks, MCP-Atlas generally exhibits the highest waste. This is because the tasks in MCP-Atlas are usually open-ended and involve a large set of unfamiliar tools and environments. Agents must first determine which tools and sources are needed before completing the tasks, which leads to more frequent detours and higher waste. Overall, the differences across benchmarks indicate that agent waste depends more on the task, the tools and the environment than on the choice of LLM. 

\subsection{Do Longer Runs Waste More?}
Intuitively, hard and exploratory tasks may result in long trajectories. One might expect longer runs to waste more in general. We test this hypothesis with two measurements. 

\begin{figure*}[!bht]
    \centering
    \includegraphics[width=\textwidth]{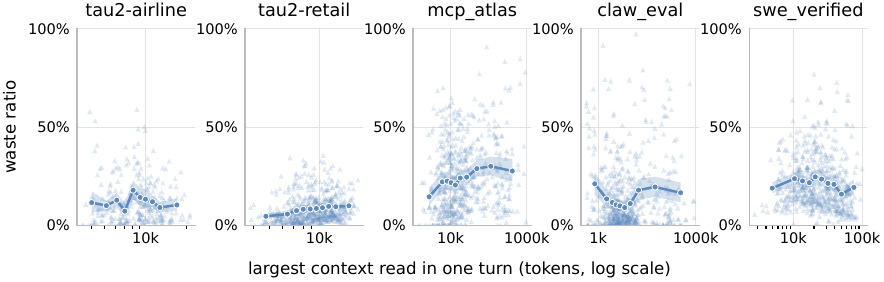}
    \caption{Waste ratio against the context length. Across benchmarks, trajectories with larger contexts can waste more, less or roughly the same share of their output. }
    \label{fig:fig6_waste_vs_context}
\end{figure*}

First, we analyze the relationship between context size and waste for each model. Figure~\ref{fig:fig6_waste_vs_context} reports the averaged waste ratio across the models against context length. Longer contexts do not necessarily lead to more waste. Across benchmarks, trajectories with larger contexts can waste more, less or roughly the same share of their output. Therefore, the relationship depends strongly on the task. Furthermore, some of the waste patterns can themselves make the context larger. For instance, detours and repeated fetches keep adding content to the trajectory, which significantly increases the context size. As a result, a larger context is often a consequence of waste, and it should not be read as its cause. For these reasons, it may be more effective to reduce waste with strategies that fit each task and agent setup, instead of uniformly forcing agents into shorter trajectories.

\begin{figure*}[t]
    \centering
    \includegraphics[width=\textwidth]{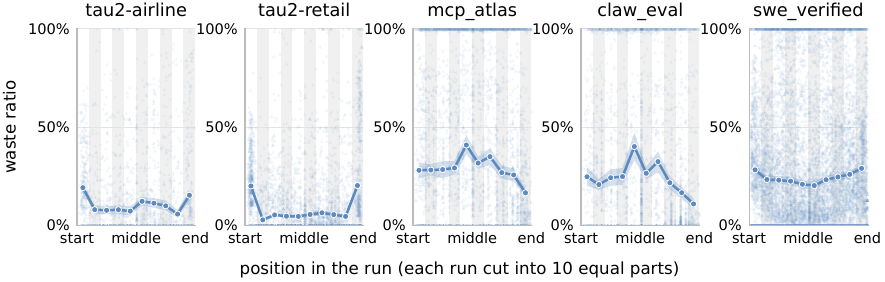}
    \caption{As the trajectory evolves, the waste ratio can either decrease or increase.}
    \label{fig:fig7_waste_along_run}
\end{figure*}

Second, the length of a trajectory can also be measured by the number of turns. Under this measure, there is no consistent pattern in how waste changes over a trajectory either, as shown in Figure~\ref{fig:fig7_waste_along_run}. Accordingly, it may be more effective to apply different strategies at different stages of a run, instead of uniformly constraining the entire trajectory.

\section{When More Reasoning Stops Paying Off}
\label{sec:when_more_reasoning_stops_paying_off}

Prior studies have shown that giving models more reasoning does not always lead to better performance~\citep{snell2025scaling,wu2025inferencescalinglawsempirical,zhou2026more,chen2026think}. This raises a natural question about what happens to the additional reasoning when increasing reasoning effort no longer improves task performance. Here, ``additional reasoning'' refers to the increase in reasoning tokens associated with higher reasoning-effort settings, rather than an extension of the same trajectory.

\begin{figure*}[!bht]
    \centering
    \includegraphics[width=\textwidth]{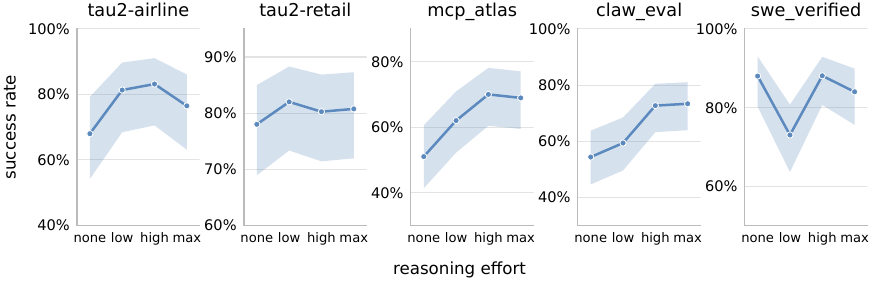}
    \caption{Success rate of Kimi-K3 across reasoning-effort settings. Increasing reasoning effort does not consistently improve task performance.}
    \label{fig:fig11_success_vs_effort}
\end{figure*}

To study the relationship between reasoning effort and token waste, we evaluate Kimi-K3 on the same four benchmarks and the same set of tasks, while varying its reasoning effort across four levels (i.e., from None, Low, High, to Max). Kimi-K3 is used for this controlled analysis, because it is the only evaluated model that exposes its complete reasoning traces. This enables direct measurement of waste in both reasoning and visible execution. In terms of task performance, the results first confirm that giving models more reasoning does not always lead to higher success rate (Figure~\ref{fig:fig11_success_vs_effort}). Additional analyses of how reasoning effort changes trajectory structure and action-level behavior are provided in Appendix~\ref{app:what_changes_when_AI_agents_reason_more}.

\subsection{Does the Additional Reasoning Become Waste?}
When more reasoning stops improving task performance, does the additional reasoning simply become waste?

\begin{figure*}[t]
    \centering
    \includegraphics[width=\textwidth]{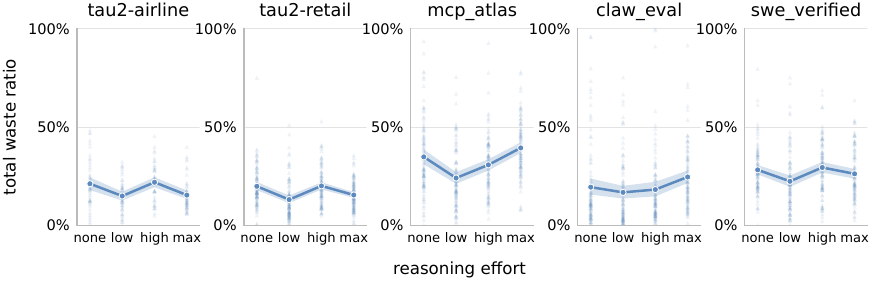}
    \caption{Overall waste ratio of Kimi-K3 across reasoning-effort settings, showing no consistent monotonic trend.}
    \label{fig:fig9_waste_vs_effort}
\end{figure*}

In our experiments, these additional reasoning tokens do not simply become waste. At Max effort, the model generates 6.6$\times$ more reasoning tokens per trajectory than at Low, but on average 66\% of this reasoning is still judged as non-wasteful. More systematically, Figure~\ref{fig:fig9_waste_vs_effort} shows that across all four effort levels and benchmarks, there exists no consistent monotonic relationship between reasoning effort and the overall waste ratio. As reasoning effort increases, waste rises on some benchmarks but falls on others, even though the agents consistently generate more reasoning. Thus, reasoning effort changes the amount of computation, but does not predict how much of this changed computation is wasted.

\begin{figure*}[!bht]
    \centering
    \includegraphics[width=\textwidth]{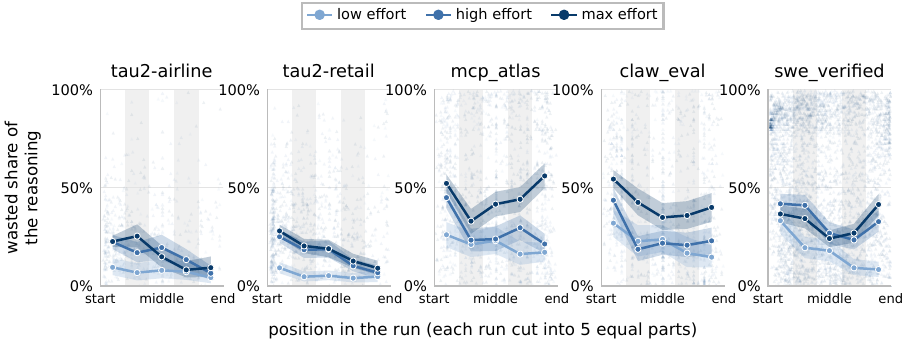}
    \caption{Waste ratio of reasoning across trajectory stages under different reasoning efforts.}
    \label{fig:fig10_reasoning_waste_by_position}
\end{figure*}

We next examine how reasoning waste changes across different stages of a trajectory. If longer reasoning simply accumulated waste, the wasted share would be expected to increase toward later stages. Instead, Figure~\ref{fig:fig10_reasoning_waste_by_position} shows that across most benchmarks and effort levels, the wasted share is highest early in the trajectory and lower toward the end. For this reason, simply cutting reasoning short may not be the best way to reduce waste.

\subsection{Where Does the Extra Reasoning Go?}
To understand where the extra reasoning goes, we separate token usage and waste in the chain of thought (i.e., thinking tokens) from those in the responses and tool calls (i.e., response tokens). 

\begin{figure*}[t]
    \centering
    \includegraphics[width=\textwidth]{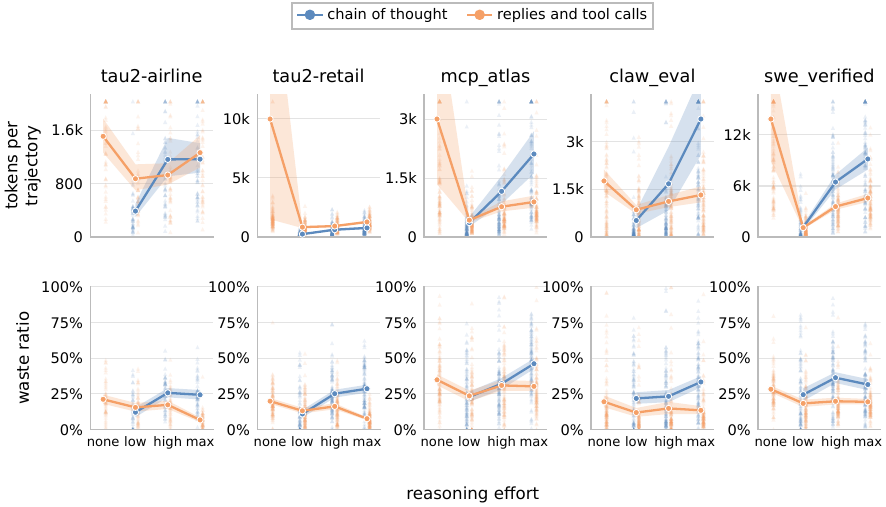}
    \caption{Token usage (top) and waste ratio (bottom) for reasoning versus visible responses and tool calls across reasoning-effort settings.}
    \label{fig:fig12_streams_vs_effort}
\end{figure*}

Figure~\ref{fig:fig12_streams_vs_effort} shows that increasing reasoning effort affects these two streams differently. Specifically, with reasoning turned off, the responses and tool calls carry a larger share of waste than at any effort level with reasoning enabled. For instance, on $\tau^2$-retail and SWE-bench Verified, the None setting produces about 12 times as many response and tool-call tokens as Low does. Accordingly, enabling reasoning generally makes the visible responses and tool calls less wasteful. However, as reasoning effort increases further from Low to High, the two streams diverge. The chain of thought grows longer and more wasteful, whereas the waste ratio of the responses and tool calls moves in no consistent direction.

The divergence between reasoning tokens and responses helps explain why the overall waste ratio shows no consistent trend, as shown in Figure~\ref{fig:fig9_waste_vs_effort}. This is because the total waste combines two sources that respond differently to reasoning effort, and their relative contributions vary across tasks. These results suggest that improving reasoning efficiency may be more effective than simply minimizing reasoning token count. More compact reasoning could preserve the benefits of reasoning while keeping the agent's responses and tool calls concise and less wasteful.

\begin{figure*}[!bht]
    \centering
    \includegraphics[width=\textwidth]{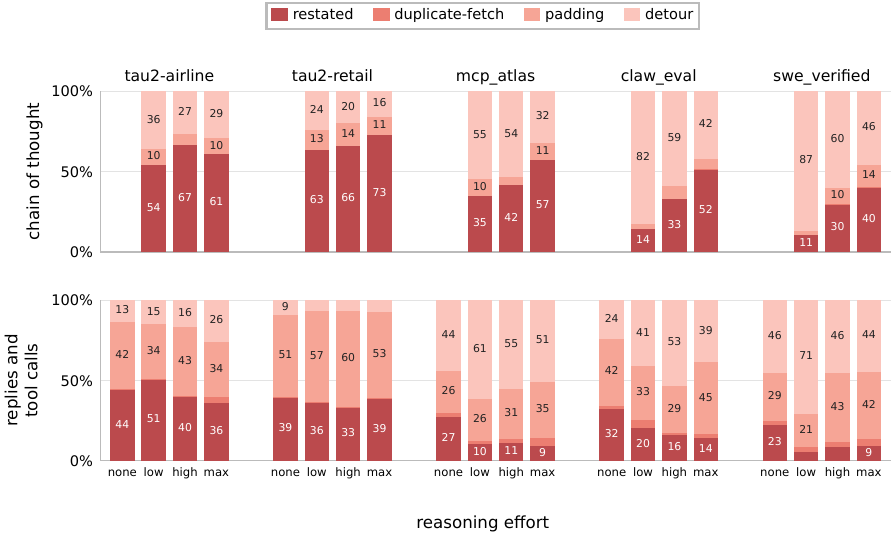}
    \caption{Waste composition across reasoning-effort settings.}
    \label{fig:fig13_waste_kinds_vs_effort}
\end{figure*}

Lastly, Figure~\ref{fig:fig13_waste_kinds_vs_effort} shows that the composition of waste also changes with reasoning effort. In general, higher effort is associated with fewer detours but more restated content in the chain of thought. In contrast, waste in responses and tool calls shows no consistent pattern across tasks.

\subsection{How Often Does Extra Reasoning Change the Outcome?}
Moving from None to Low slightly improves performance while using 1.48 million fewer tokens. From Low to High, an additional 0.99 million tokens yields a clear improvement, increasing success rate from 69.7\% to 78.4\%. However, moving from High to Max consumes another 0.80 million tokens without measurable performance improvement. 

\begin{figure*}[!bht]
    \centering
    \includegraphics[width=0.6\textwidth]{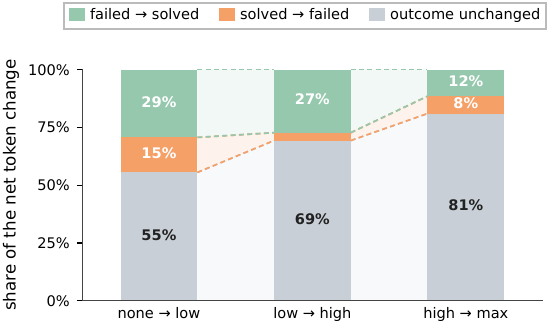}
    \caption{Change in output tokens between adjacent reasoning-effort levels, paired task by task and grouped by task-outcome changes. None $\rightarrow$ Low reduces token usage, while higher effort levels increase it.}
    \label{fig:fig15_where_the_tokens_go}
\end{figure*}

Figure~\ref{fig:fig15_where_the_tokens_go} further shows how often the additional reasoning changes task outcomes. From Low to High, 69\% of the additional tokens are spent on tasks whose outcomes remain unchanged, while 27\% are associated with failures that become successful. From High to Max, 81\% of the additional tokens are spent on tasks with unchanged outcomes, while only 12\% are associated with failures that Max fixes. Among the tasks solved by both High and Max, Max costs 1.62 times as much, while the overall waste ratio changes only slightly from 23.7\% to 24.9\%. This observation indicates that, at high reasoning effort, the agent spends much more computation on longer paths through the same tasks, even when those paths do not change the outcome, while the share of wasted content changes little.
\section{How Does Harness Design Shape Token Waste?}
\label{sec:same_model_different_agent}
An agent harness is the software layer that orchestrates an LLM to use tools, interact with the environment, and manage the agent’s execution loop. The experiments in the previous sections primarily use the native harnesses provided by each benchmark. Here, with the same LLM and the same reasoning effort, we compare different harnesses and investigate how harness design affects agent behavior and token waste.

We run Kimi-K3 at the same reasoning effort across five harnesses on MCP-Atlas and SWE-bench Verified. Under this controlled setting, changing the harness markedly alters token usage, action patterns, and workflow structure. However, differences in waste are not consistently related to broad harness properties, such as tool count or the supported action space. Some waste can be traced directly to harness implementation and interaction design, while other patterns persist across harnesses and instead appear to depend on the model and task.

For instance, several harnesses launch SWE-bench Verified tasks in the wrong Python environment, causing import errors and repeated retries across many trajectories. This waste can be directly attributed to the flawed harness design. In comparison, there exists another recurring source of waste that appears at nearly the same rate across all harnesses. That is, Kimi-K3 tends to write out what it thinks is the project’s original fix from memory and later abandons it. Therefore, this pattern is more likely to come from the model and task, rather than from the harness design.

\subsection{Does Less Waste Mean Better Performance?}
\begin{figure*}[!bht]
    \centering
    \includegraphics[width=0.8\textwidth]{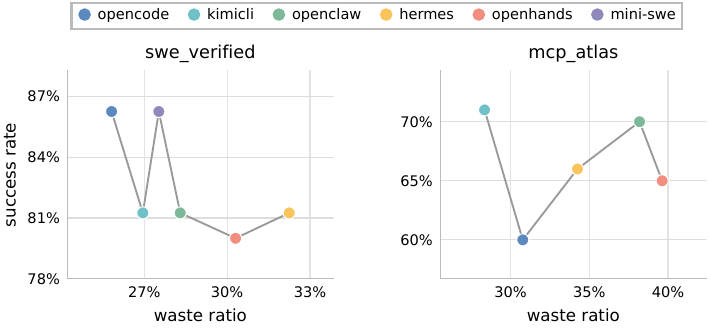}
    \caption{Success rate versus token-waste ratio across agent harnesses with the same Kimi-K3 model and reasoning effort. Lower waste does not consistently correspond to higher task performance.}
    \label{fig:fig19_waste_vs_success_by_harness}
\end{figure*}

\begin{figure*}[!bht]
    \centering
    \includegraphics[width=\textwidth]{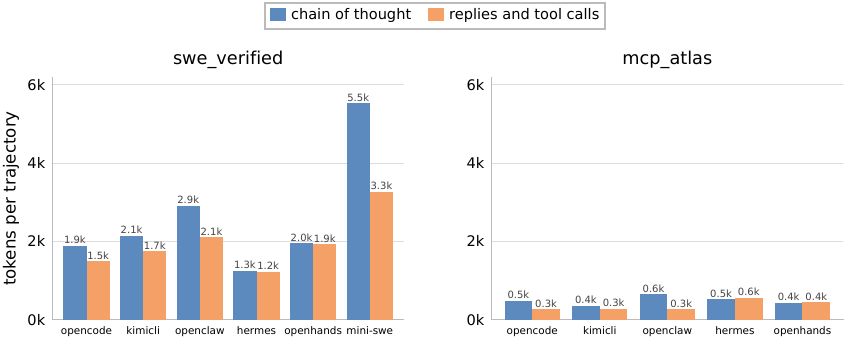}
    \caption{Median token usage per trajectory across agent harnesses, separated into chain-of-thought and visible responses and tool calls.}
    \label{fig:fig20_tokens_by_harness}
\end{figure*}

Across both SWE-bench Verified and MCP-Atlas, harnesses with less waste do not consistently achieve better performance (Figure~\ref{fig:fig19_waste_vs_success_by_harness}). Some harnesses produce noticeably less waste while achieving similar task performance to more wasteful ones. Thus, differences in token waste do not directly explain differences in task performance across harnesses. In fact, what changes more clearly is how the agent executes the task. Figure~\ref{fig:fig20_tokens_by_harness} shows that Kimi-K3 can produce more than four times as much reasoning in one harness as in another, even on the same tasks and at the same reasoning effort.

\begin{figure*}[t]
    \centering
    \includegraphics[width=\textwidth]{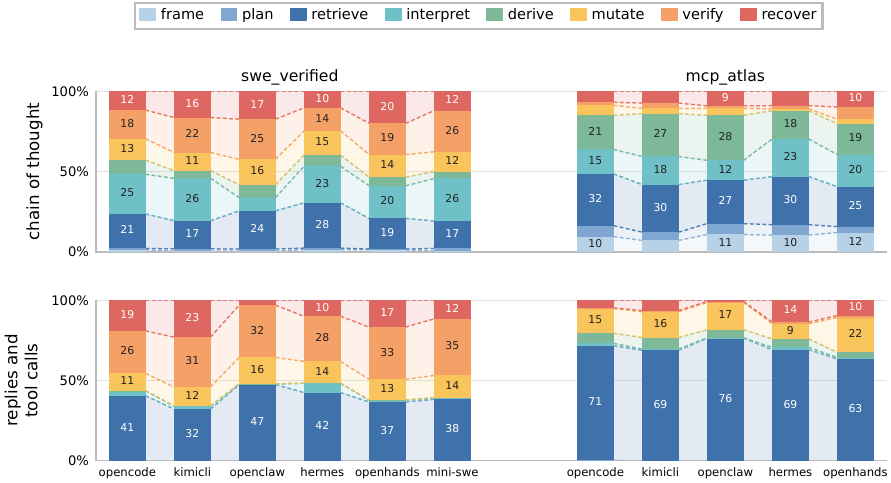}
    \caption{Distribution of action categories across agent harnesses, shown separately for chain-of-thought (top) and visible responses and tool calls (bottom).}
    \label{fig:fig21_action_mix_by_harness}
    \vspace{-0.4cm}
\end{figure*}

Harnesses also induce different action distributions as shown in Figure~\ref{fig:fig21_action_mix_by_harness}. For example, in visible execution, the share of \textsc{Recover} differs by about 20 percentage points across harnesses on SWE-bench Verified, while the share of \textsc{Retrieve} differs by more than 10 percentage points on MCP-Atlas. Thus, the harness changes how much the agent writes as well as what the agent spends its turns doing.

However, detailed action patterns do not transfer consistently between tasks. For example, Kimi Code CLI produces the highest \textsc{Recover} share among the harnesses on SWE-bench Verified, but only the third highest on MCP-Atlas. In comparison, some whole-trajectory properties appear more stable. For instance, on MCP-Atlas, roughly 10\% of OpenHands trajectories account for 73\% of all OpenHands turns, while on SWE-bench Verified, long trajectories account for about 40\% of its turns. More broadly, harnesses that produce more repetitive or unusually long trajectories tend to show similar tendencies across the two benchmarks. Overall, the detailed action mix is highly task-dependent, whereas broader properties such as repetition and runaway behavior appear more stable across tasks.

\subsection{Where Does Harness-Induced Waste Come from?}

\begin{figure*}[!bht]
    \centering
    \includegraphics[width=\textwidth]{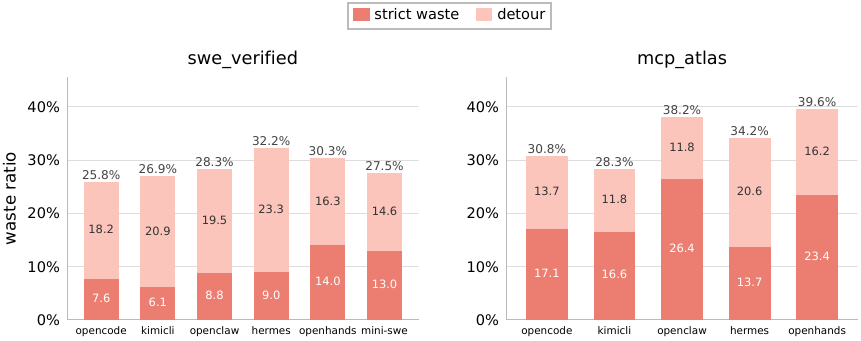}
    \caption{Token-waste ratio across agent harnesses, decomposed into strict waste and detours on SWE-bench Verified and MCP-Atlas.}
    \label{fig:fig25_waste_split_by_harness}
\end{figure*}

We next ask which differences in waste can actually be attributed to the harness. Figure~\ref{fig:fig25_waste_split_by_harness} shows that different harnesses exhibit different mixtures of strict waste and detours. On SWE-bench Verified, Hermes has the highest overall waste ratio, but much of this waste comes from detours. Mini-SWE has a lower overall waste ratio, but it has one of the highest strict-waste ratios. Thus, looking only at total waste can obscure why a harness is wasteful.

\begin{figure*}[!bht]
    \centering
    \includegraphics[width=\textwidth]{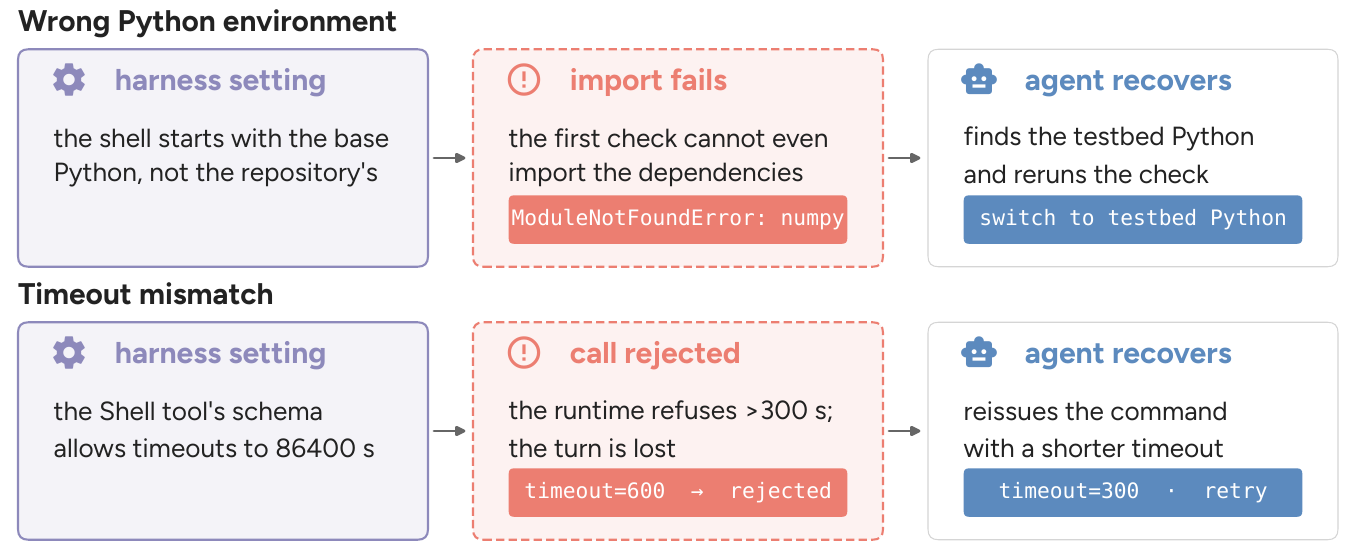}
    \caption{Two exemplar sources of harness-induced waste.}
    \label{fig:fig28_why_the_harness_wastes}
\end{figure*}

Furthermore, we inspect the harness implementations and trace several sources of waste directly to implementation or control-loop design. Figure~\ref{fig:fig28_why_the_harness_wastes} highlights two recurring sources of harness-induced waste. One clear source is that OpenCode, Kimi Code CLI, and OpenClaw launch the shell in an incorrect Python environment on SWE-bench Verified, causing import failures and repeated retries across many trajectories. Kimi Code CLI exposes a timeout value that its own runtime rejects, forcing the model to repeat the call. These examples show that harness-induced waste often comes from how the harness connects the model to the environment and structures the interaction, rather than from the model selecting the wrong type of work.

\begin{figure*}[!bht]
    \centering

    \begin{subfigure}[t]{0.45\textwidth}
        \centering
        \includegraphics[width=\linewidth]{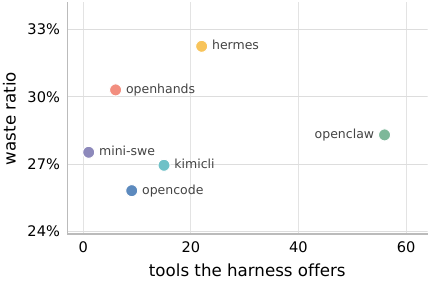}
        \caption{Tool count versus overall waste ratio across harnesses on SWE-bench Verified.}
        \label{fig:fig26a_tool_count_vs_waste}
    \end{subfigure}
    \hfill
    \begin{subfigure}[t]{0.45\textwidth}
        \centering
        \includegraphics[width=\linewidth]{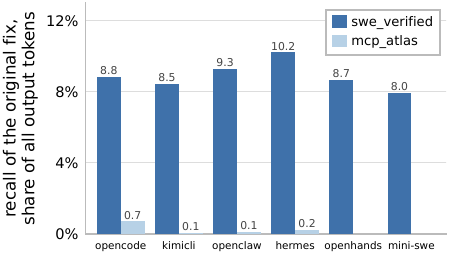}
        \caption{Recurring original-fix detour across harnesses and tasks.}
        \label{fig:fig26d_original_fix}
    \end{subfigure}

    \caption{
    Diagnostics of harness-related token waste across complementary aspects of harness design and agent execution.
    }
    \label{fig:fig26_what_explains_harness_waste}
\end{figure*}

Surprisingly, several seemingly obvious harness properties do not consistently explain the observed differences in waste. For instance, as shown in Figure~\ref{fig:fig26a_tool_count_vs_waste}, harnesses that expose more tools are not necessarily more wasteful. OpenClaw exposes many more tools than the other harnesses, yet exhibits only moderate waste. Moreover, not all waste changes with the harness. Figure~\ref{fig:fig26d_original_fix} demonstrates that on SWE-bench Verified, there is a recurring detour in which Kimi-K3 writes out what it believes to be the project's original fix and later abandons it. This pattern appears at nearly the same rate across harnesses, but is almost absent on MCP-Atlas. Since this pattern persists when the harness changes but largely disappears when the task changes, it is better characterized as a model-task effect than as a harness effect. Such waste may be difficult to eliminate through harness design alone.

\subsection{Harness-Environment Mismatch Can Create Avoidable Waste}
\begin{figure*}[!bht]
    \centering
    \includegraphics[width=\textwidth]{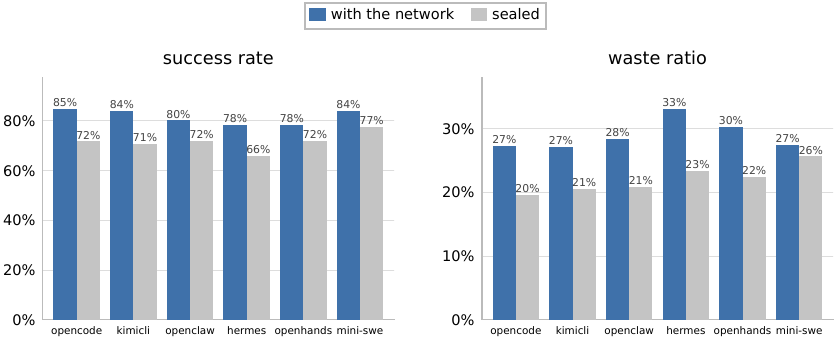}
    \caption{Task success rate (left) and token-waste ratio (right) across agent harnesses on SWE-bench Verified, with and without network access.}
    \label{fig:fig27_network_ablation}
\end{figure*}

Finally, we test a direct intervention on the harness by removing the agent's access to the internet. As shown in Figure~\ref{fig:fig27_network_ablation}, this experiment reveals another source of inefficiency when the harness fails to adapt to constraints in the execution environment.

When network access is removed, agents often continue retrying installations or probing alternative ways to reach unavailable resources. These actions may be locally reasonable from the model's perspective, but once the required capability is unavailable, continuing the trajectory is unlikely to improve the outcome.

Interestingly, these network-disabled runs produce substantially more output while exhibiting lower measured waste ratios. This happens because the waste judge sees many retries as locally reasonable rather than removable. The result exposes a broader form of harness inefficiency. An agent can spend many additional tokens pursuing a task that has effectively become infeasible, even when those individual actions are not classified as token waste.

This suggests an additional principle for harness design beyond reducing redundant text or tool calls. A harness can detect environmental constraints, communicate them clearly to the model, and terminate execution when further progress is unlikely or impossible. In such cases, the task may still fail, but the harness can avoid spending large amounts of computation on a trajectory whose outcome is unlikely to change.
\section{Related Work: From Token Consumption to Token Waste}
\label{sec:related_work}
\paragraph{Agent cost and efficiency.}
The growing cost of LLM agents has motivated a broad line of work on agent efficiency. 
Recent studies mainly characterize agent execution via aggregate quantities such as token consumption, monetary cost, latency, interaction steps, and resource budgets~\citep{bai2026ai,lin2026bagen,wang2025efficient,chen2026depo,su2026beyond}. For example, in~\cite{bai2026ai}, the author study end-to-end token consumption across models and tasks in agentic coding tasks, showing substantial variation across models, tasks, and repeated executions. BAGEN~\citep{lin2026bagen} treats budget as an active control signal and investigates whether agents can estimate their remaining resource requirements during execution. 
DEPO~\citep{chen2026depo} further distinguishes efficiency at the interaction-step and trajectory levels, and optimizes both tokens per step and the number of steps required to complete a task. Beyond token counts, \citet{su2026beyond} propose a hardware-aware metric for tool-integrated reasoning that captures inference costs not reflected by token or tool-call counts alone. These studies establish resource consumption as an important dimension of agent efficiency. 
Our work focuses on a complementary perspective. Specifically, total consumption measures how much computation an agent uses, whereas \textbf{token waste} asks how much of that computation is unnecessary and removable. Under our framework, a trajectory can be long but contain little removable waste, whereas a shorter trajectory may still contain substantial redundancy.

\paragraph{Redundancy and waste within agent trajectories.}
Beyond aggregate cost, several recent studies examine redundancy inside agent trajectories. 
RedundancyBench~\citep{hu2026redundant} formulates redundant-step detection and asks whether individual turns in a successful trajectory contribute meaningfully to task completion. Some other related work identifies repeated or unnecessary computation at other granularities, including duplicated information in multi-agent communication~\citep{wang2025agenttaxo}, redundant messages~\citep{zhang2025cut}, redundant agents and communication links~\citep{wang2025agentdropout,zhang2026safesieve}, and repeated tool-execution patterns~\citep{abuzakuk2026optimizing}. These studies are closely related to our notion of token waste, but primarily characterize redundancy at the level of complete steps, inter-agent communication, or tool sequences. 
In contrast, we identify removable \textbf{spans} within reasoning, visible responses, and tool interactions. We further distinguish directly avoidable waste from exploratory detours and analyze how these waste patterns vary across models, tasks, reasoning settings, and agent harnesses.

\paragraph{Reasoning efficiency and test-time compute.}
Another line of work studies the trade-off between additional inference-time reasoning and task performance. 
Inference-scaling studies examine how additional test-time compute should be allocated to maximize performance under a fixed budget~\citep{wu2025inferencescalinglawsempirical}. 
More recent work shows that longer reasoning is not always better: excessive test-time computation can exhibit diminishing returns or even reduce accuracy through overthinking~\citep{zhou2026more}, while raw reasoning length itself may be a poor proxy for the amount of meaningful reasoning performed~\citep{chen2026think}.
Our analysis complements this literature by separating \emph{reasoning length} from \emph{reasoning waste}. 
In particular, reasoning can stop improving the final task outcome without becoming entirely removable. We further study how reasoning effort changes agent actions and workflow structure, and show that waste can evolve differently in internal reasoning and visible execution. Thus, performance saturation, longer reasoning, and token waste capture related but distinct notions of agent efficiency.

\paragraph{Agent harness and workflow design.}
The execution system surrounding an LLM can fundamentally affect agent performance and efficiency. 
Harness-Bench~\citep{yao2026harness} evaluates model-harness combinations under shared task environments and finds significant variation in completion, efficiency, and failure behavior across configurations. Life-Harness~\citep{xu2026adapting} improves frozen LLM agents by adapting the runtime interface between the model and environment. Agentic Harness Engineering~\citep{lin2026agentic} automatically evolves harness components such as tools, middleware, and memory and reports improvements in both task performance and token usage. These studies establish that harness design can affect agent performance and aggregate efficiency.  Our work instead focuses on \emph{waste attribution}. When two harnesses produce different token-waste patterns under the same model and reasoning setting, which differences can actually be attributed to the harness?  We analyze how harnesses alter action distributions, workflow stability, and reasoning frequency, and distinguish waste introduced by harness implementation and interaction design from patterns that persist across harnesses and are better explained by the model and task.
\section{Conclusion}
\label{sec:conclusion}

In this work, we introduce a unified framework for measuring and analyzing token waste in AI agents. The framework traces heterogeneous agent executions and normalizes them into a common representation. It follows a minimalism principle: content is considered waste if it can be removed without eliminating useful work from the trajectory. Based on this definition, the framework identifies removable content in reasoning, visible responses, and tool interactions. It then quantifies the corresponding token and monetary waste. We validate this measurement through systematic intervention experiments. The results show that the identified waste can generally be removed with limited impact on subsequent agent behavior and final task performance. This makes it possible to distinguish token consumption from token waste, rather than treating all generated tokens as equally necessary. Across the collected trajectories, roughly 18\% of output tokens are identified as waste, and for more than 45\% of tasks, the wasted output of one agent alone exceeds the entire output of another agent that successfully solves the same task. In online pruning experiments, removing 41-46\% of the model's reasoning still largely preserves task performance across SWE-bench Verified and MCP-Atlas.

We evaluate the framework across six frontier LLMs and multiple agentic benchmarks to compare how token waste varies across models and tasks. Then, a controlled reasoning study varies the reasoning effort of Kimi-K3 to examine how additional reasoning changes token usage and waste. Finally, we compare multiple agent harnesses using the same model and tasks to examine how harness design affects token waste. Across the collected trajectories, roughly 18\% of output tokens are identified as waste. Waste also varies across tasks and environments. Reasoning and visible execution exhibit different waste patterns, and harness design can introduce distinct sources of inefficiency. Together, these findings show that aggregate token usage alone is not enough to understand agent efficiency. By making unnecessary computation measurable and attributable, our framework provides diagnostic signals for improving models, reasoning strategies, and agent harnesses. 

\section*{Limitations}
\label{sec:limitation}
Our study has several limitations. First, token waste is defined by whether content can be removed without eliminating useful work. Although we use an LLM-as-a-Judge and intervention-based validation to make this measurement systematic, some cases remain ambiguous, especially detours that are unnecessary in hindsight but reasonable when attempted. More broadly, our metric focuses on \emph{removable content} and may miss other forms of trajectory-level inefficiency, such as repeatedly pursuing locally reasonable actions after environmental constraints have made the task effectively infeasible.

Second, complete reasoning traces are not exposed by all model providers. Our controlled analysis of reasoning effort therefore focuses on Kimi-K3, whose full reasoning channel is observable. For models that expose only summaries or no reasoning text, waste measurements are limited to observable trajectory content, so conclusions about internal reasoning may not generalize to all models.

Third, our experiments cover a finite set of frontier models, benchmarks, and agent harnesses. These settings span multiple task domains and execution systems but do not capture the full diversity of agent architectures, tools, environments, or interaction patterns. Harness implementations also evolve rapidly, so specific implementation-level waste patterns may change over time.


\bibliography{colm2026_conference}

@inproceedings{
    jimenez2024swebench,
    title={{SWE}-bench: Can Language Models Resolve Real-world Github Issues?},
    author={Carlos E Jimenez and John Yang and Alexander Wettig and Shunyu Yao and Kexin Pei and Ofir Press and Karthik R Narasimhan},
    booktitle={The Twelfth International Conference on Learning Representations},
    year={2024},
    url={https://openreview.net/forum?id=VTF8yNQM66}
}

@inproceedings{
  wang2025openhands,
  title={OpenHands: An Open Platform for {AI} Software Developers as Generalist Agents},
  author={Xingyao Wang and Boxuan Li and Yufan Song and Frank F. Xu and Xiangru Tang and Mingchen Zhuge and Jiayi Pan and Yueqi Song and Bowen Li and Jaskirat Singh and Hoang H. Tran and Fuqiang Li and Ren Ma and Mingzhang Zheng and Bill Qian and Yanjun Shao and Niklas Muennighoff and Yizhe Zhang and Binyuan Hui and Junyang Lin and Robert Brennan and Hao Peng and Heng Ji and Graham Neubig},
  booktitle={The Thirteenth International Conference on Learning Representations},
  year={2025},
  url={https://openreview.net/forum?id=OJd3ayDDoF}
}

@misc{openai2025codex,
  author       = {{OpenAI}},
  title        = {Introducing Codex},
  year         = {2025},
  url          = {https://openai.com/blog/introducing-codex}
}

@misc{wu2025inferencescalinglawsempirical,
      title={Inference Scaling Laws: An Empirical Analysis of Compute-Optimal Inference for Problem-Solving with Language Models}, 
      author={Yangzhen Wu and Zhiqing Sun and Shanda Li and Sean Welleck and Yiming Yang},
      year={2025},
      eprint={2408.00724},
      archivePrefix={arXiv},
      primaryClass={cs.AI},
      url={https://arxiv.org/abs/2408.00724}, 
}

@inproceedings{wang2025agenttaxo,
  title={AgentTaxo: Dissecting and Benchmarking Token Distribution of LLM Multi-Agent Systems},
  author={Wang, Qian and Tang, Zhenheng and Jiang, Zichen and Chen, Nuo and Wang, Tianyu and He, Bingsheng},
  year         = {2025},
  booktitle={ICLR 2025 Workshop on Foundation Models in the Wild}
}

@inproceedings{yao2022react,
  title={React: Synergizing reasoning and acting in language models},
  author={Yao, Shunyu and Zhao, Jeffrey and Yu, Dian and Du, Nan and Shafran, Izhak and Narasimhan, Karthik R and Cao, Yuan},
  booktitle={The eleventh international conference on learning representations},
  year={2022}
}

@inproceedings{
snell2025scaling,
title={Scaling {LLM} Test-Time Compute Optimally Can be More Effective than Scaling Parameters for Reasoning},
author={Charlie Victor Snell and Jaehoon Lee and Kelvin Xu and Aviral Kumar},
booktitle={The Thirteenth International Conference on Learning Representations},
year={2025},
url={https://openreview.net/forum?id=4FWAwZtd2n}
}

@article{wang2025efficient,
  title={Efficient agents: Building effective agents while reducing cost},
  author={Wang, Ningning and Hu, Xavier and Liu, Pai and Zhu, He and Hou, Yue and Huang, Heyuan and Zhang, Shengyu and Yang, Jian and Liu, Jiaheng and Zhang, Ge and others},
  journal={arXiv preprint arXiv:2508.02694},
  year={2025}
}

@inproceedings{liu2024agentbench,
  title={Agentbench: Evaluating llms as agents},
  author={Liu, Xiao and Yu, Hao and Zhang, Hanchen and Xu, Yifan and Lei, Xuanyu and Lai, Hanyu and Gu, Yu and Ding, Hangliang and Men, Kaiwen and Yang, Kejuan and others},
  booktitle={International Conference on Learning Representations},
  volume={2024},
  pages={52989--53046},
  year={2024}
}

@article{yang2024swe,
  title={Swe-agent: Agent-computer interfaces enable automated software engineering},
  author={Yang, John and Jimenez, Carlos and Wettig, Alexander and Lieret, Kilian and Yao, Shunyu and Narasimhan, Karthik and Press, Ofir},
  journal={Advances in Neural Information Processing Systems},
  volume={37},
  pages={50528--50652},
  year={2024}
}

@article{packer2023memgpt,
  title={Memgpt: Towards llms as operating systems},
  author={Packer, Charles and Wooders, Sarah and Lin, Kevin and Fang, Vivian and Patil, Shishir G and Stoica, Ion and Gonzalez, Joseph E},
  journal={arXiv preprint arXiv:2310.08560},
  year={2023}
}

@article{wu2023autogen,
  title={Autogen: Enabling next-gen llm applications via multi-agent conversation},
  author={Wu, Qingyun and Bansal, Gagan and Zhang, Jieyu and Wu, Yiran and Li, Beibin and Zhu, Erkang and Jiang, Li and Zhang, Xiaoyun and Zhang, Shaokun and Liu, Jiale and others},
  journal={arXiv preprint arXiv:2308.08155},
  year={2023}
}

@article{ruan2026beyond,
  title={Beyond Prompts: Measuring and Optimizing LLM Tool-Agent Harnesses},
  author={Ruan, Haibo and Chen, Wenjie and Tu, Pei-fen and Abbasi, Usman and Hesch, Joel and others},
  journal={arXiv preprint arXiv:2609.05736},
  year={2026}
}

@article{xu2026adapting,
  title={Adapting the interface, not the model: Runtime harness adaptation for deterministic llm agents},
  author={Xu, Tianshi and Wen, Huifeng and Li, Meng},
  journal={arXiv preprint arXiv:2605.22166},
  year={2026}
}

@article{yao2026harness,
  title={Harness-Bench: Measuring harness effects across models in realistic agent workflows},
  author={Yao, Yilun and Tan, Xinyu and Liu, Chao-Hsuan and Li, Yaoming and Wang, Zhengyang and Yu, Wenhan and Tan, Zhewen and Tian, Yuxuan and Zhao, Guangxiang and Sun, Lin and others},
  journal={arXiv preprint arXiv:2605.27922},
  year={2026}
}

@article{luo2025ultrahorizon,
  title={Ultrahorizon: Benchmarking agent capabilities in ultra long-horizon scenarios},
  author={Luo, Haotian and Zhang, Huaisong and Zhang, Xuelin and Wang, Haoyu and Qin, Zeyu and Lu, Wenjie and Ma, Guozheng and He, Haiying and Xie, Yingsha and Zhou, Qiyang and others},
  journal={arXiv preprint arXiv:2509.21766},
  year={2025}
}

@article{bai2026ai,
  title={How do ai agents spend your money? analyzing and predicting token consumption in agentic coding tasks},
  author={Bai, Longju and Huang, Zhemin and Wang, Xingyao and Sun, Jiao and Mihalcea, Rada and Brynjolfsson, Erik and Pentland, Alex and Pei, Jiaxin},
  journal={arXiv preprint arXiv:2604.22750},
  year={2026}
}

@article{satish2026beyond,
  title={Beyond Token Savings: A Systematic Study of Context Compression in LLM Agents},
  author={Satish, Ritul and Sinha, Prasoon and Kawada, Akiho and Yadwadkar, Neeraja J},
  journal={arXiv preprint arXiv:2609.32961},
  year={2026}
}

@article{lin2026bagen,
  title={BAGEN: Are LLM Agents Budget-Aware?},
  author={Lin, Yuxiang and Wang, Zihan and Liu, Mengyang and Shan, Yuxuan and Bai, Longju and Zhang, Junyao and Jin, Xing and Chen, Boshan and Su, Jinyan and Wang, Xingyao and others},
  journal={arXiv preprint arXiv:2606.00198},
  year={2026}
}

@misc{barres2025tau2,
      title={$\tau^2$-Bench: Evaluating Conversational Agents in a Dual-Control Environment}, 
      author={Victor Barres and Honghua Dong and Soham Ray and Xujie Si and Karthik Narasimhan},
      year={2025},
      eprint={2506.07982},
      archivePrefix={arXiv},
      primaryClass={cs.AI},
      url={https://arxiv.org/abs/2506.07982}, 
}

@misc{ye2026clawevaltrustworthyevaluationautonomous,
      title={Claw-Eval: Towards Trustworthy Evaluation of Autonomous Agents}, 
      author={Bowen Ye and Rang Li and Qibin Yang and Yuanxin Liu and Linli Yao and Hanglong Lv and Zhihui Xie and Chenxin An and Lei Li and Lingpeng Kong and Qi Liu and Zhifang Sui and Tong Yang},
      year={2026},
      eprint={2604.06132},
      archivePrefix={arXiv},
      primaryClass={cs.AI},
      url={https://arxiv.org/abs/2604.06132}, 
}

@article{bandi2026mcp,
  title={Mcp-atlas: A large-scale benchmark for tool-use competency with real mcp servers},
  author={Bandi, Chaithanya and Dumitru, Razvan-Gabriel and Hertzberg, Ben and Agarwal, Divyansh and Boo, Geobio and Polakam, Tejas and Hassaan, Sami and Da, Jeff and Kim, HiJae and Gupta, Vipul and others},
  journal={arXiv preprint arXiv:2602.00933},
  year={2026}
}

@misc{google2026gemini31pro,
  author       = {{Google DeepMind}},
  title        = {Gemini 3.1 Pro},
  year         = {2026},
  month        = feb,
  howpublished = {Model Card},
  url          = {https://deepmind.google/models/model-cards/gemini-3-1-pro/},
  note         = {Published February 19, 2026}
}

@misc{openai2026gpt56,
  author       = {{OpenAI}},
  title        = {{GPT-5.6 System Card}},
  year         = {2026},
  month        = jul,
  howpublished = {System Card},
  url          = {https://deploymentsafety.openai.com/gpt-5-6},
  note         = {Published July 9, 2026}
}

@misc{openai2026gpt6astra,
  author       = {{OpenAI}},
  title        = {{GPT-6 Astra System Card}},
  year         = {2026},
  month        = sep,
  howpublished = {System Card},
  url          = {https://deploymentsafety.openai.com/gpt-6-astra},
  note         = {Published September 3, 2026}
}

@misc{anthropic2026sonnet5,
  author       = {{Anthropic}},
  title        = {{Claude Sonnet 5 System Card}},
  year         = {2026},
  month        = jun,
  howpublished = {System Card},
  url          = {https://www.anthropic.com/claude-sonnet-5-system-card}
}

@misc{anthropic2026fable51,
  author       = {{Anthropic}},
  title        = {{Claude Fable 5.1 and Claude Mythos 5.1 System Card}},
  year         = {2026},
  month        = sep,
  howpublished = {System Card},
  url          = {https://www.anthropic.com/claude-fable-5-1-mythos-5-1-system-card}
}

@article{team2026kimi,
  title={Kimi k3: Open frontier intelligence},
  author={Team, Kimi and Bai, Tongtong and Bai, Yifan and Bao, Yiping and Cai, Jianfeng and Cai, Xinyuan and Cao, Peizhou and Cao, Yuxuan and Chai, Ziwei and Charles, Y and others},
  journal={arXiv preprint arXiv:2607.24653},
  year={2026}
}

@misc{opencode2026,
  author       = {{OpenCode Team}},
  title        = {OpenCode: The Open Source Coding Agent},
  year         = {2026},
  howpublished = {Software},
  url          = {https://github.com/anomalyco/opencode}
}

@misc{openclaw2026,
  author       = {{OpenClaw Team}},
  title        = {OpenClaw},
  year         = {2026},
  howpublished = {Software},
  url          = {https://github.com/openclaw/openclaw}
}

@misc{nous2026hermes,
  author       = {{Nous Research}},
  title        = {Hermes Agent},
  year         = {2026},
  month        = feb,
  howpublished = {Software},
  url          = {https://nousresearch.com/releases},
  note         = {Released February 25, 2026}
}

@misc{moonshot2026kimicode,
  author       = {{Moonshot AI}},
  title        = {Kimi Code CLI},
  year         = {2026},
  howpublished = {Software},
  url          = {https://github.com/MoonshotAI/kimi-code},
  note         = {AI coding agent for the terminal}
}

@inproceedings{chen2026depo,
  title={Depo: Dual-efficiency preference optimization for llm agents},
  author={Chen, Sirui and Zhao, Mengshi and Xu, Lei and Zhao, Yuying and Zhu, Beier and Zhang, Hanwang and Zhao, Shengjie and Lu, Chaochao},
  booktitle={Proceedings of the AAAI Conference on Artificial Intelligence},
  volume={40},
  number={36},
  pages={30279--30287},
  year={2026}
}

@inproceedings{su2026beyond,
  title={Beyond Accuracy: Unveiling Inefficiency Patterns in Tool-Integrated Reasoning},
  author={Su, Qisheng and Huang, Shiting and Fang, Zhen and Chen, Ziyan and Chen, Zehui and Zhao, Feng},
  booktitle={Proceedings of the 64th Annual Meeting of the Association for Computational Linguistics (Volume 1: Long Papers)},
  pages={7456--7477},
  year={2026}
}

@article{hu2026redundant,
  title={Redundant or Necessary? A Benchmark for Detecting Redundant Steps in Agent Trajectories},
  author={Hu, Minyang and Yang, Bo and Zhou, Zhinuo and Liang, Jiachen and Jiahao, Guo and Yin, Yiyang and Han, Xiongwei},
  journal={arXiv preprint arXiv:2605.29893},
  year={2026}
}

@article{abuzakuk2026optimizing,
  title={Optimizing agentic workflows using meta-tools},
  author={Abuzakuk, Sami and Kermarrec, Anne-Marie and Sharma, Rishi and Veski, Rasmus Moorits and de Vos, Martijn},
  journal={arXiv preprint arXiv:2601.22037},
  year={2026}
}

@article{chen2026think,
  title={Think deep, not just long: Measuring LLM reasoning effort via deep-thinking tokens},
  author={Chen, Wei-Lin and Peng, Liqian and Tan, Tian and Zhao, Chao and Chen, Blake JianHang and Lin, Ziqian and Go, Alec and Meng, Yu},
  journal={arXiv preprint arXiv:2602.13517},
  year={2026}
}

@inproceedings{zhou2026more,
  title={When more thinking hurts: Overthinking in LLM test-time compute scaling},
  author={Zhou, Shu and Ling, Rui and Chen, Junan and Wang, Xin and Fan, Tao and Wang, Hao},
  booktitle={Findings of the Association for Computational Linguistics: ACL 2026},
  pages={23967--23977},
  year={2026}
}

@article{lin2026agentic,
  title={Agentic harness engineering: Observability-driven automatic evolution of coding-agent harnesses},
  author={Lin, Jiahang and Liu, Shichun and Pan, Chengjun and Lin, Lizhi and Dou, Shihan and Xi, Zhiheng and Huang, Xuanjing and Yan, Hang and Han, Zhenhua and Gui, Tao and others},
  journal={arXiv preprint arXiv:2604.25850},
  year={2026}
}

@inproceedings{zhang2025cut,
  title={Cut the crap: An economical communication pipeline for llm-based multi-agent systems},
  author={Zhang, Guibin and Yue, Yanwei and Li, Zhixun and Yun, Sukwon and Wan, Guancheng and Wang, Kun and Cheng, Dawei and Yu, Jeffrey and Chen, Tianlong},
  booktitle={International Conference on Learning Representations},
  volume={2025},
  pages={75389--75428},
  year={2025}
}

@inproceedings{wang2025agentdropout,
  title={Agentdropout: Dynamic agent elimination for token-efficient and high-performance llm-based multi-agent collaboration},
  author={Wang, Zhexuan and Wang, Yutong and Liu, Xuebo and Ding, Liang and Zhang, Miao and Liu, Jie and Zhang, Min},
  booktitle={Proceedings of the 63rd Annual Meeting of the Association for Computational Linguistics (Volume 1: Long Papers)},
  pages={24013--24035},
  year={2025}
}

@inproceedings{zhang2026safesieve,
  title={Safesieve: From heuristics to experience in progressive pruning for llm-based multi-agent communication},
  author={Zhang, Ruijia and Zhao, Xinyan and Wang, Ruixiang and Chen, Sigen and Zhang, Guibin and Zhang, An and Wang, Kun and Wen, Qingsong},
  booktitle={Proceedings of the AAAI Conference on Artificial Intelligence},
  volume={40},
  number={35},
  pages={29892--29900},
  year={2026}
}

@inproceedings{wang2025steca,
  title={Steca: Step-level trajectory calibration for llm agent learning},
  author={Wang, Hanlin and Wang, Jian and Leong, Chak Tou and Li, Wenjie},
  booktitle={Findings of the Association for Computational Linguistics: ACL 2025},
  pages={11597--11614},
  year={2025}
}
\bibliographystyle{colm2026_conference}
\newpage

\appendix
\section{What Changes When AI Agents Reason More}
\label{app:what_changes_when_AI_agents_reason_more}
Higher reasoning effort generally leads a model to generate more reasoning tokens, but token count is not the only thing that changes. In this section, we examine how reasoning effort affects trajectory structure, action-level behavior, and the distribution of token waste.

\subsection{More Reasoning Can Mean Deeper Steps or More Steps}
\begin{figure*}[!bht]
    \centering
    \includegraphics[width=\textwidth]{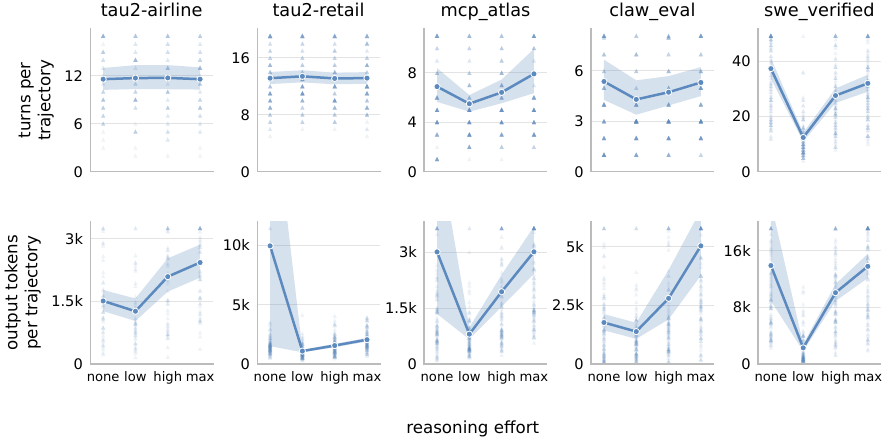}
    \caption{Average number of turns (top) and output tokens (bottom) per trajectory across reasoning-effort settings.}
    \label{fig:fig14_run_shape_vs_effort}
\end{figure*}

Increasing reasoning effort can change the shape of an agent trajectory in different ways. As shown in Figure~\ref{fig:fig14_run_shape_vs_effort}, on $\tau^2$-bench, the number of turns remains nearly unchanged as reasoning effort increases, while the number of output tokens grows. Therefore, the additional computation primarily comes from more reasoning within roughly the same sequence of actions. In contrast, on SWE-bench Verified, the average number of turns increases with reasoning effort, resulting in longer trajectories.

These results reveal two distinct ways in which an agent uses a larger reasoning budget. The agent can reason more deeply between roughly the same actions, or it can take more actions and extend the trajectory. This distinction is important because an increase in reasoning tokens does not necessarily mean that the model is simply thinking longer about the same steps. Higher reasoning effort can instead change the structure of the execution itself, potentially creating new places for both useful work and waste to occur.

\subsection{How Does Reasoning Effort Change the Agent's Actions?}
Changes in trajectory length tell us \emph{how} an agent uses additional computation, but not \emph{what} that computation is used for. The next step is to examine whether reasoning effort also changes the agent's behavior within the trajectory.

To characterize the agent's behavior, we separately classify the model's reasoning and its visible execution for each turn into eight action categories: \textbf{Frame}, \textbf{Plan}, \textbf{Retrieve}, \textbf{Interpret}, \textbf{Derive}, \textbf{Mutate}, \textbf{Verify}, and \textbf{Recover}:
\begin{itemize}
    \item Frame: Understand what the task is and what constraints apply, such as policy, permission or format requirements.
    \item Plan: Deciding the overall approach or sequence of steps for the task. Choosing only the next action does not count as planning.
    \item Retrieve: Looking up or reading information from the environment without changing anything.
    \item Interpret: Making sense of information returned by a tool before deciding what to do next.
    \item Derive: Reaching a conclusion from information already available, such as through arithmetic, counting or inference, without looking anything up.
    \item Mutate: Creating the deliverable itself, such as writing code, making an edit, or composing the final response. Drafting the content counts even before it is applied or sent.
    \item Verify: Checking whether a result matches expectations, before any problem has been identified. Running a test is a verify. Responding to a failed test is Recover.
    \item Recover: Responding when something has gone wrong, such as diagnosing the failure, backing out of an approach or switching to a different one.
\end{itemize}

\begin{figure*}[!bht]
    \centering
    \includegraphics[width=\textwidth]{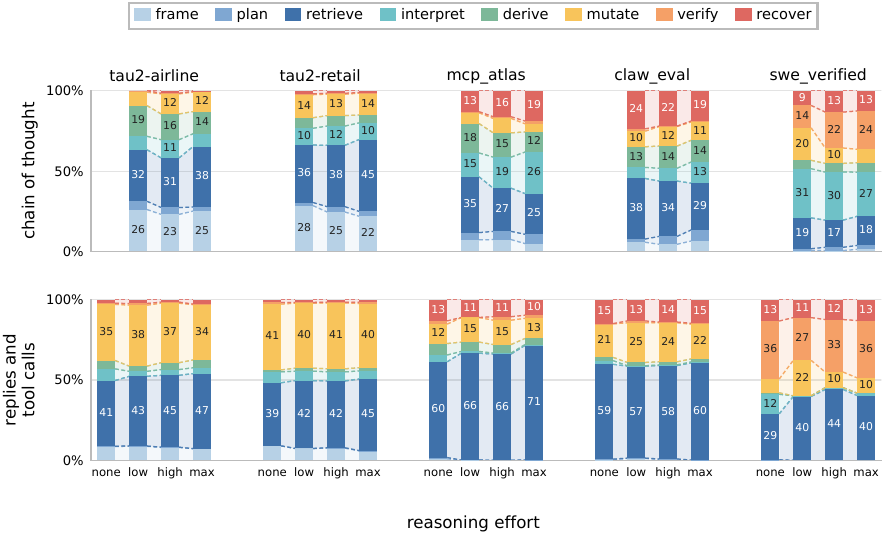}
    \caption{Distribution of action categories in chain-of-thought (top) and visible responses and tool calls (bottom) across reasoning-effort settings.}
    \label{fig:fig16_action_mix_vs_effort}
\end{figure*}

We then compare the distribution of these actions across reasoning-effort levels to understand how additional reasoning changes the work performed by the agent. Figure~\ref{fig:fig16_action_mix_vs_effort} demonstrates that increasing reasoning effort changes what the agent uses the additional reasoning for. For instance, as reasoning effort increases, the agent can spend less reasoning on “Derive” (i.e., reaching a conclusion from the information already available) but more on interacting with the environment (e.g., Retrieve, Interpret, Verify, Recover).

In addition, a large change in the agent’s visible actions between None and Low. When reasoning is switched from None to Low, the visible ratio of Interpret drops consistently across all five task sets. With reasoning disabled, Interpret reaches its highest ratio in the visible actions. In contrast, once reasoning is enabled, much less interpretation remains in the visible stream. This suggests that the reasoning channel becomes an important place where agents interpret tool results.

Lastly, higher reasoning effort does not simply increase reasoning within a single action. The additional reasoning budget is distributed across different action categories, and this distribution is highly task-dependent. For instance, on SWE-bench Verified the largest increase is in Verify, on the $\tau^2$-bench task sets it is Retrieve, and on MCP Atlas and Claw-Eval it is Interpret. One outlier is Derive, whose ratio either falls or stays nearly unchanged across all five task sets.

\subsection{Does the Action Become More Wasteful?}
\begin{figure*}[!bht]
    \centering
    \includegraphics[width=\textwidth]{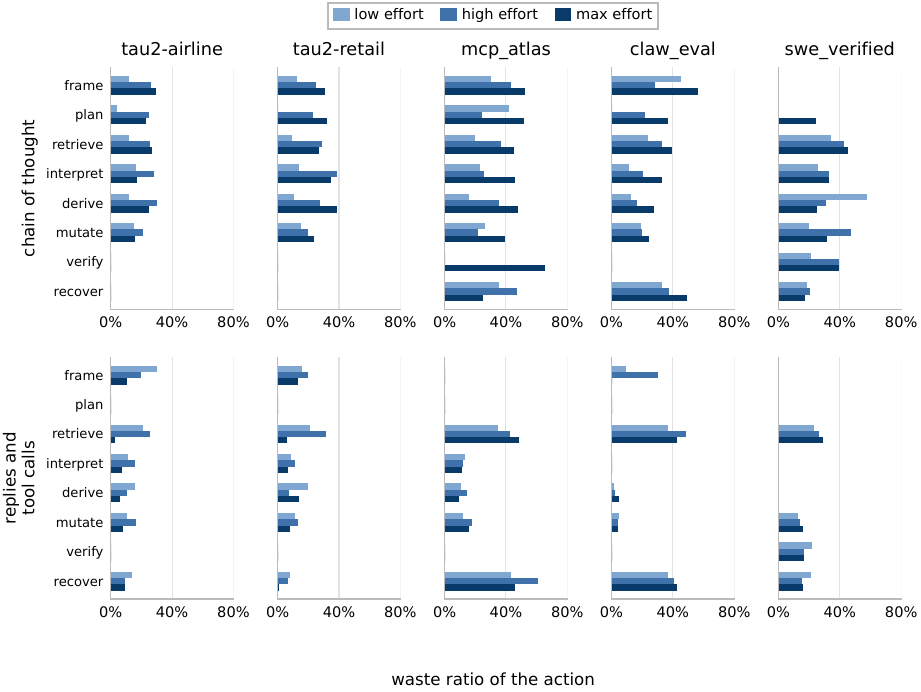}
    \caption{Waste ratio within each action category across reasoning-effort settings, shown separately for chain-of-thought (top) and visible responses and tool calls (bottom).}
    \label{fig:fig18_waste_by_action_vs_effort}
\end{figure*}

Reasoning effort also impacts how wasteful the same action becomes. As shown in Figure~\ref{fig:fig18_waste_by_action_vs_effort}, this effect differs sharply between internal reasoning and visible execution. The relative mix of visible actions changes comparatively little, whereas the waste within those actions can still change largely.

Within the reasoning tokens, waste ratio generally increases within the same action categories as reasoning effort increases. Visible responses and tool calls show the opposite pattern. In the visible content, the same actions generally become cleaner and less wasteful. Thus, higher reasoning effort can increase waste in internal reasoning while reducing it in visible execution. This pattern explains why overall waste does not move monotonically with reasoning effort.

Moreover, this distinction also suggests that ``use fewer tokens'' is too coarse as an optimization objective. Reasoning and visible execution may require separate token-efficiency controls, potentially adapted to different stages of the workflow.

Overall, more reasoning changes more than the number of tokens an agent generates. Higher reasoning effort changes what the agent does, how the workflow is structured, and where waste appears within that workflow.

\section{LLMaJ Efficiency Judge: System Prompt}
\label{app:llmaj-efficiency}

Token-waste identification uses an LLM-as-a-Judge to audit each complete agent trajectory. The judge receives every turn in order, including the preceding user or tool input and the agent's reasoning, visible response, and tool calls when available. Following the system prompt below, it identifies removable spans, assigns each span to a waste category, and records which later parts of the trajectory depend on the remaining content. Note that the identified unread tokens are too few to be significant and therefore, are ignored in this work.

\begin{tcolorbox}[
    colback=yellow!5,
    colframe=yellow!60!black,
    breakable,
    left=2pt, right=2pt, top=2pt, bottom=2pt,
    before upper=\setlength{\parskip}{6pt}
]

\paragraph{System Role}

You audit how an AI agent spent its output tokens.

You get one complete trajectory: every turn it took, in order. Each turn shows what the agent had just received -- the user's message, or the previous tool's result -- and then up to three streams of its own output:
\begin{itemize}
  \item \texttt{reasoning}: the model's private thinking for this turn
  \item \texttt{visible}: the text the user would see
  \item \texttt{action}: the tool calls the turn sent, with their arguments
\end{itemize}

Received text is marked ``from the user'' or ``from a tool''. Neither is audited; both are input, and you are auditing what the agent produced. They are shown because they are what makes ``was this needed'' answerable, and in two directions: a call that repeats an earlier one is milder when the user asked for it in between, and reasoning that reads the user's own words back is not new content however well phrased.

Your job for each stream is to mark what could be removed.

\paragraph{Marking}

Quote every stretch of the stream that could be deleted without changing the outcome. Not a few examples -- every one. A stream that is load-bearing overall still contains sentences that carry nothing, and those are what you are looking for.

Each quote needs a kind:
\begin{itemize}
  \item \texttt{restated}: says again what is already in the context -- the user's own words, an earlier tool result, or the agent's own earlier turn
  \item \texttt{duplicate-fetch}: a call whose result is the same as one already in the context, with nothing but tool output in between -- nobody asked for it and nothing could have changed
  \item \texttt{padding}: words with no content: filler openers, commentary on style, sign-offs, restated plans
  \item \texttt{detour}: a step that need not have been taken at all -- a trial, an exploration, a route later abandoned
  \item \texttt{unread}: produced, and then nothing ever reads it
\end{itemize}

Copy the text word for word: quotes are located in the source, and one that cannot be found is recorded as a fabrication rather than a finding.

Up to 40 quotes per stream. If you reach that many, quote the longest stretches -- a cap that binds is reported, so do not compress your answer to fit.

\paragraph{What You Keep}

Whatever you do not mark is your claim that those characters were load-bearing, and it is read literally. Two fields state it:
\begin{itemize}
  \item \texttt{kept}: one sentence naming what the unmarked remainder does. If you marked the whole stream, say plainly that nothing is left.
  \item \texttt{depended\_on\_by}: the round number of the work that uses that remainder -- the edit it enables, the answer it supplies a fact to, the call whose result gets used. For reasoning this is often the same round, since a turn's thinking feeds that turn's own action. An array; empty when nothing uses any of it.
\end{itemize}

You are not asked to grade the stream. Necessity is what is left after you mark, so there is no label to fall back on -- a stream you mark nothing in is one you are asserting was necessary character for character.

\paragraph{Rules}

\begin{itemize}
  \item If nothing at all uses a stream, mark the whole thing and leave \texttt{depended\_on\_by} empty. That is a finding, not a failure to answer.
  \item Reading the request back is \texttt{restated}, even when compressed and even when it is the opening move. So is quoting the standing instructions back, when they are shown to you. Summarising in order to reach a conclusion is not: the conclusion is new, so mark the summary and keep the conclusion.
  \item Applying a rule is not repeating it. ``Basic economy is not refundable'' copied out of the standing instructions is \texttt{restated}; ``this booking is basic economy, so it is not refundable'' joins the rule to a fact and the result was in neither. Mark the part that was already written down and keep the part that was not.
  \item A call repeated with a different result is flagged for you in a note, and it is yours to rule on. If the agent changed something and is confirming the change, the re-check is necessary and you mark nothing. If it was trying a variation that did not help, it is \texttt{detour}. The note tells you the result differed; it does not tell you which of those happened.
  \item Getting the same information twice: the first time stands, the repeats are \texttt{duplicate-fetch}. Re-running a test after an edit is not a repeat -- the earlier result no longer holds. Re-reading a file nobody touched is.
  \item Proximity is not necessity. A comment inside a necessary patch, a preamble before a necessary edit, a sign-off after a correct answer: the work around them stays and they go.
  \item A stream cannot outrank what it serves. Reasoning that sets up a step you marked \texttt{detour} is itself at most \texttt{detour}.
  \item Judge against the task, not against whether the run succeeded -- you are not told whether it did.
  \item A repeat the user asked for is still a repeat, but a milder one. Only a repeat with nothing but tool output in between is \texttt{duplicate-fetch}. When the customer asked for a double-check in between, the call is \texttt{detour}: it could have come out differently and it complied with a request -- but the same bytes re-entered the context and the agent already had them.
  \item Quote word for word. Quotes are located in the source text, and one that is not found there is recorded as a fabrication.
\end{itemize}

\paragraph{Alignment}

Every stream's entry must carry \texttt{opening}: the first 25 characters of the stream you are auditing, copied exactly. Its only job is to prove the entry is filed against the right turn, so copy it rather than summarising it.

\paragraph{Output}

A JSON object with a \texttt{turns} array, one entry per turn, in the order given. Each turn lists only the streams it actually has:

\begin{Verbatim}[breaklines, breakanywhere, fontsize=\small]
{"turns": [
  {"round": <copied from the input>,
    "streams": [
      {"stream": "reasoning",
        "opening": "<first 25 characters, word for word>",
        "spans": [{"quote": "<word for word>", "kind": "restated",
                   "why": "<one sentence>"},
                  {"quote": "<word for word>", "kind": "padding",
                   "why": "<one sentence>"}],
        "kept": "<one sentence>",
        "depended_on_by": [<round numbers>]},
      {"stream": "action", "opening": "...", "spans": [],
        "kept": "...", "depended_on_by": []}
    ]}
]}
\end{Verbatim}

Copy every round number from the input; they are checked.

\end{tcolorbox}

\end{document}